\UseRawInputEncoding
\documentclass[runningheads]{llncs}

\usepackage{eccv}

\usepackage{eccvabbrv}

\usepackage{graphicx}
\usepackage{booktabs}
\usepackage[utf8]{inputenc}
\usepackage{arydshln}
\usepackage{colortbl}
\usepackage{xcolor}
\usepackage{tikz}
\usetikzlibrary{calc}
\usetikzlibrary{spy}

\pgfdeclarelayer{background}
\pgfsetlayers{background,main}

\definecolor{best}{RGB}{255, 153, 153}    
\definecolor{second}{RGB}{255, 204, 153}  
\definecolor{third}{RGB}{255, 255, 153}   

\newcommand{\colorfirst}{255, 153, 153}
\newcommand{\colorsecond}{255, 204, 153}
\newcommand{\colorthird}{255, 255, 153}
\definecolor{colorfirst}{RGB}{255, 153, 153}
\definecolor{colorsecond}{RGB}{255, 204, 153}
\definecolor{colorthird}{RGB}{255, 255, 153}

\usepackage[accsupp]{axessibility}  

\usepackage{xcolor} 

\definecolor{colorfirst}{RGB}{255,180,180}
\definecolor{colorsecond}{RGB}{255,220,170}
\definecolor{colorthird}{RGB}{255,245,170}

\usepackage{hyperref}

\usepackage{orcidlink}

\usepackage{xcolor}

\begin{document}


\title{GrowFields: \\ Compositional 4D Neural Fields for \\ Topology-Changing Plant Growth}
\titlerunning{GrowFields}

\author{Joaquin Gajardo\inst{1}\orcidlink{0000-0001-9144-4384} \and
Michele Volpi\inst{2}\orcidlink{0000-0003-2771-0750} \and
Marko Mihajlovic\inst{1}\orcidlink{0000-0001-6305-3896} \and
Siyu Tang\inst{1}\orcidlink{0000-0002-1015-4770} \and
Lukas Roth\inst{1,3}\orcidlink{0000-0003-1435-9535} \and
Sergey Prokudin\inst{1}\orcidlink{0000-0001-6501-8234}}

\authorrunning{J.~Gajardo et al.}

\institute{
ETH Zürich, Switzerland \and
Swiss Data Science Center, Switzerland \and
Martin-Luther-Universität Halle-Wittenberg, Germany \\
\url{https://joaquin-gajardo.github.io/growfields}
}

\maketitle

\begin{abstract}

Quantifying plant growth dynamics from sparse longitudinal 3D observations is fundamental for agriculture and plant sciences.
Yet, plants pose unique challenges: they undergo intricate non-rigid deformations, exhibit changing topology as new organs emerge, and often lack explicit temporal correspondences between consecutive data acquisitions due to newly formed tissue.
Methods designed for general scenes struggle to model topology changes and asynchronous organ growth characteristic of plants.
To address these challenges, we introduce GrowFields, a compositional dynamic neural field representation for organ-aware 4D plant growth modelling from point cloud time series.
Our approach decomposes a plant into its constituent organs and aligns each organ into its own canonical coordinate frame, isolating intrinsic growth patterns from global plant motion. 
We then learn a shared continuous neural deformation field that models temporal dynamics across all organs, conditioned on learnable per-organ latent codes capturing organ identity and growth characteristics.
The resulting modular yet unified representation naturally accommodates the asynchronous development of plant organs while remaining grounded in the practical setting of organ-level plant tracking.
We evaluate GrowFields on growth sequences from four plant species, assessing geometric fitting and organ tracking accuracy using manually annotated leaf-tip trajectories.
Results demonstrate consistent improvements in spatial precision, temporal coherence, and morphological fidelity over a range of existing representations.

\end{abstract}    
\section{Introduction}
\label{sec:intro}

\begin{figure}[t]
  \centering
   \includegraphics[width=\linewidth]{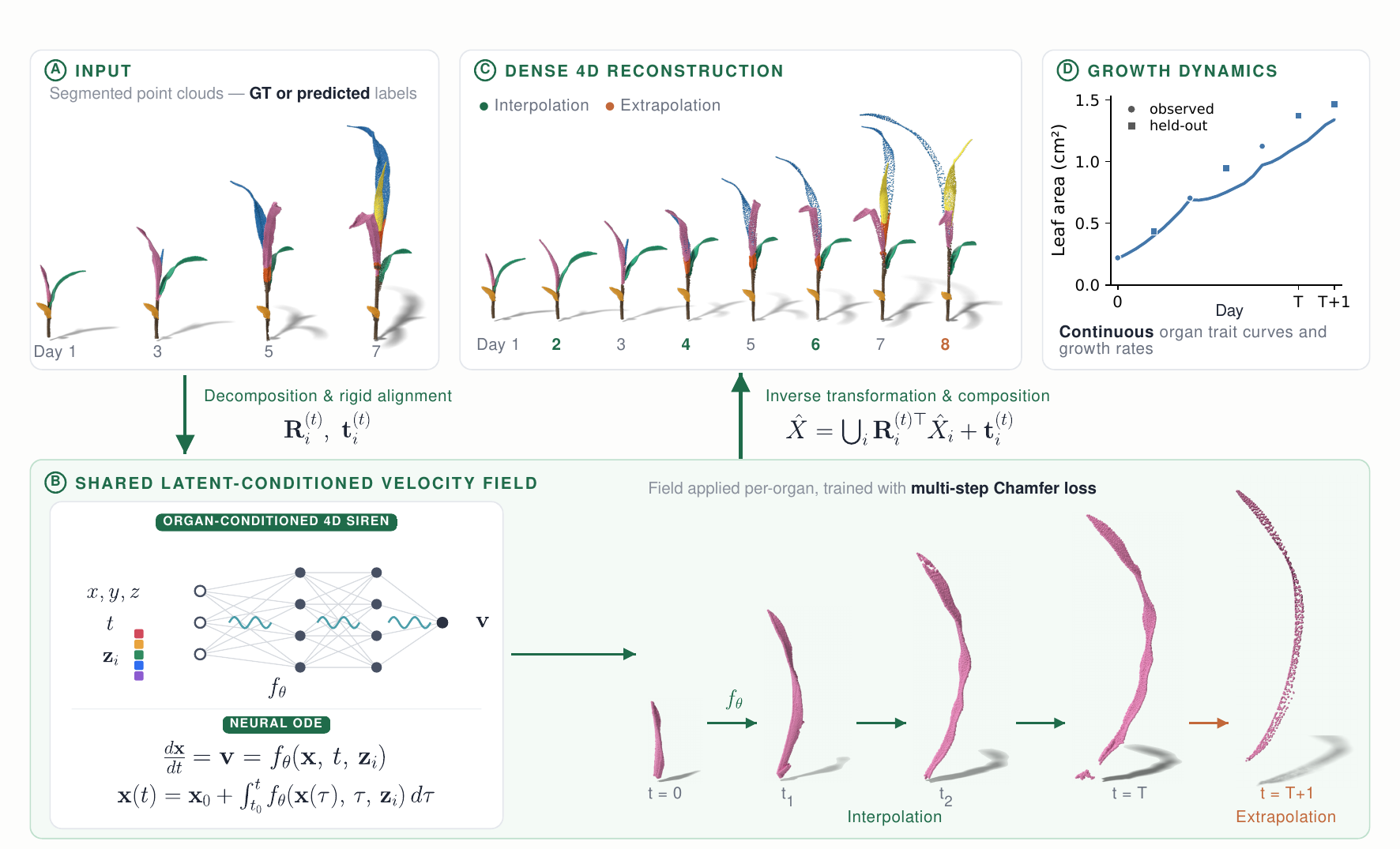}
   \caption{\textbf{Overview of GrowFields.}
Given a time series of segmented plant point clouds, the plant is decomposed into organs and each organ is aligned to a canonical coordinate frame \textbf{(A)}. 
For an initial point $\mathbf{x}_0$ belonging to organ $i$, its trajectory over time is predicted by integrating a shared neural deformation field $f_\theta$ \textbf{(B)}, conditioned on a learnable organ-specific latent code $\mathbf{z}_i$:
$\mathbf{x}(t,\mathbf{z}_i,\mathbf{x}_0)=\mathbf{x}_0+\int_0^t f_\theta(\mathbf{x}(\tau),\tau,\mathbf{z}_i)\,d\tau$.
The predicted organ motions are recomposed in the global coordinate frame to produce a temporally coherent 4D reconstruction of the full plant \textbf{(C)}.
The SIREN-based field~\cite{Sitzmann_2020_Neurips} produces smooth trajectories for all points, modelling continuous organ growth dynamics \textbf{(D)}.
}
   \label{fig:teaser}
\end{figure}

Understanding and quantifying plant growth dynamics is central to agriculture and plant sciences.
Plants are complex, dynamic, multi-scale organisms whose response to environmental conditions can be partly characterised through the temporal dynamics of observable organ-level traits, such as the timing of organ emergence and leaf expansion rates \cite{granier_multi-scale_2009}.
While 3D sensing technologies and plant modelling methods have advanced significantly, enabling detailed reconstructions of plant geometry \cite{okura_3d_2022, Cheng_2025_ICCV, yang_gaussianplant_2025}, the temporal dimension of plant growth remains underexplored.
Modelling plants in 4D from sparse temporal observations could enable continuous measurement of organ-specific growth rates, early stress detection, and quantitative analysis of plant development, providing a powerful new tool for plant scientists and agronomists \cite{furbank_phenomics_2011}. 

Faithfully capturing these dynamics at the organ level remains challenging.
Plants undergo complex non-rigid deformations, exhibit changing topology as new organs appear and old ones decay, and data acquisition is often sparse and noisy.
Furthermore, newly formed tissue has no direct temporal correspondence, breaking assumptions typically made in non-rigid registration and dynamic scene reconstruction.

General-purpose 4D reconstruction techniques, based on temporal Signed Distance Functions (SDFs) \cite{Wang_2025_ICCV, Sang_2025_CVPR}, or flow- \cite{niemeyer2019occupancy} and point-based dynamic models \cite{Prokudin_2023_ICCV}, usually assume globally consistent topology or smoothly varying deformation fields.
Recent work such as GrowFlow~\cite{luo_grow_2026} models plant growth through a continuous deformation field learned in reverse time to mitigate the appearance of new geometry.
While effective for modelling smooth plant deformation, such global formulations do not explicitly model organ identity or 
asynchronous organ development, making it difficult to represent organ emergence, large non-rigid growth, and long-term organ-level dynamics. 
Other plant-specific methods leverage domain-specific cues such as organ segmentation \cite{magistri_segmentation-based_2020} or skeleton extraction \cite{chebrolu_registration_2021, pan_multi-scale_2021, zhang_spatio-temporal_2023, khanam_riemannian_2024}, but often rely on multi-stage pipelines and manually engineered matching strategies that may not generalise across species or sensing conditions. 
Moreover, most existing approaches address reconstruction, tracking, and dynamic modelling in isolation, and therefore struggle to jointly capture continuous plant deformation, organ identity, and topology changes under realistic temporally sparse acquisition settings.

Motivated by the observation that plant organs grow asynchronously and follow distinct deformation patterns, we propose \emph{GrowFields}, a part-based dynamic neural field framework for continuous 4D plant reconstruction from longitudinal point clouds (\autoref{fig:teaser}).
Instead of modelling plants as a single deforming structure, our approach leverages their natural modular organisation: the plant is decomposed into organs whose geometry and motion can be analysed independently while remaining part of a coherent whole.
By representing organ dynamics in canonical reference frames and learning a shared neural model of growth conditioned on learnable latent codes representing organ identity, the framework captures common geometric and temporal patterns while allowing individual organs to evolve according to their own trajectories.
Operating directly on segmented point clouds aligns with common phenotyping workflows \cite{li_survey_2025, Zhang_Gajardo_2025_CVPR}, while the compositional neural field formulation provides a flexible framework for learning nonlinear plant dynamics at the organ level. 

To rigorously assess the effectiveness of GrowFields, we conduct a comprehensive evaluation across multiple plant species exhibiting diverse morphologies, growth patterns, and environmental conditions.
Our experimental setup includes newly curated leaf-tip annotations on relevant public datasets, enabling quantitative assessment of temporal coherence, organ-level tracking, and geometric stability.
This application-grounded evaluation design, combined with physiologically relevant metrics, provides a thorough empirical validation of the proposed framework and establishes a challenging benchmark for future research.
We further probe the method's practicality through experiments on robustness to noise and missing data, temporal extrapolation, and automatically segmented and tracked organ labels.

To summarise, our main contributions are:
\begin{itemize}
\item We introduce \textbf{GrowFields}, a compositional neural field framework for modelling topology-changing plant growth from sparse longitudinal point clouds as a modular dynamical system.
It combines organ canonicalisation with a shared continuous velocity field conditioned on per-organ latent codes, effectively capturing large non-rigid growth, continuous deformation, and asynchronous organ development.
\item We present an \textbf{open-source benchmark and evaluation protocol} for 4D plant growth modelling based on annotated leaf-tip trajectories, enabling quantitative evaluation of temporal coherence and organ-level tracking accuracy.
\end{itemize}

\section{Related work}
\label{sec:related_work}

Understanding plant development from 3D observations requires representations able to capture geometric structure, temporal evolution, and organ-level organisation under real-world data acquisition constraints.
Thus, we review prior work on plant reconstruction, dynamic plant models, organ segmentation and tracking, and dynamic and compositional neural representations.

\subsubsection{Static plant models.}

Plant modelling has long been studied in computer graphics and plant sciences \cite{okura_3d_2022, Deussen_1998, walter_2012_botany}.
Recent approaches reconstruct plant geometry directly from images or point clouds~\cite{luo_leaffit_2026, liu_neural_2025, Cheng_2025_ICCV, Yang_2025_ICCV, yang_gaussianplant_2025}.
Notably, 3D Gaussian Splatting (3DGS)~\cite{kerbl3Dgaussians} has recently become popular for high-fidelity plant reconstruction~\cite{li_survey_2025, ojo_splanting_2024, stuart_high-fidelity_2025, shen_plantgaussian_2025, shen_biomass_2025, an_agrigaussian_2026} and yield-component organs~\cite{Zhang_Gajardo_2025_CVPR, ding_grpe_2025, jiang_cotton3dgaussians_2025}, but does not explicitly encode plant structure.
Structured alternatives have started to emerge, including neural leaf models~\cite{Yang_2025_ICCV}, plant mesh reconstruction methods~\cite{luo_leaffit_2026}, and parametric plant representations such as Demeter~\cite{Cheng_2025_ICCV} and GaussianPlant~\cite{yang_gaussianplant_2025}.
While these methods achieve high-quality 3D geometry, they remain restricted to static scenes.
We instead model plant development over time, enabling the characterisation of dynamic plant behaviour and the computation of continuous growth variables.

\vspace{-4mm}
\subsubsection{Dynamic plant models.}

Early work analysed temporal plant point clouds to detect growth events or track organs across time~\cite{li_analyzing_2013, paproki_2012}.
Subsequent approaches focused on skeleton- or graph-based registration~\cite{chebrolu_registration_2021, pan_multi-scale_2021, zhang_spatio-temporal_2023, meyer_cherrypicker_2023}, typically extracting plant topology from point clouds via Laplacian contraction~\cite{Cao_2010_laplacian} or $L_1$-medial contraction~\cite{huang_2013_acm}.
In parallel, visual SLAM~\cite{dong_4d_2017} or 3DGS~\cite{adebola2025growsplat} have also been used to visualise temporally consistent appearance changes.
Some work has also sought to interpolate registered 3D scans by simple spherical linear interpolation (SLERP)~\cite{pan_multi-scale_2021, zhang_spatio-temporal_2023} or optimal transport~\cite{golla_temporal_2020}.
More recently, GrowFlow~\cite{luo_grow_2026} models plant growth in reverse time by combining 3DGS with a HexPlane~\cite{cao_hexplane_2023} deformation field guided by Neural Ordinary Differential Equations (Neural ODEs)~\cite{chen_neural_2018}, achieving improved interpolation results over general-purpose 4D Gaussian Splatting (4DGS)~\cite{Wu_2024_CVPR}.
However, evaluation is limited to synthetic plants and real plants without large organ-level changes.
Despite these advances, most existing approaches do not support organ-level modelling and tracking, which is essential for phenotyping and plant analysis~\cite{furbank_phenomics_2011}.
In contrast, we learn compositional dynamic neural fields from segmented longitudinal data, enabling structure-aware modelling of asynchronous multi-organ plant development.

\vspace{-4mm}
\subsubsection{Organ segmentation and tracking.}

Organ-level decomposition is central to plant phenotyping.
Methods include direct 3D instance segmentation of point clouds \cite{song_comprehensive_2025, Cheng_2025_ICCV, du_pst_2023, li_psegnet_2022} using annotated datasets~\cite{zhu_crops3d_2024, mertoglu_2023, schunck_pheno4d_2021}, and image-based segmentation projected into 3D~\cite{van_marrewijk_3d_2025, yang_plantsegnerf_2026}.
Recent approaches leverage vision foundation models~\cite{xing_zero-shot_2025, singh_few-shot_2025} or specialised segmentation networks~\cite{chen_gmt_2025}, often combined with neural rendering frameworks such as PlantSegNeRF~\cite{yang_plantsegnerf_2026}.
Segmentation has also enabled organ tracking and emergence detection, using bipartite matching~\cite{li_trackplant3d_2024, freeman_transformer-based_2025}, task-specific detection frameworks~\cite{li_3d-nod_2025}, or visual place recognition and feature matching~\cite{riccardi_fruit_2023, fusaro_horticultural_2026, lobefaro_spatio-temporal_2025}. We build upon these foundations to learn plant growth dynamics from sparse temporal data, motivating the use of dynamic neural representations capable of capturing large, non-linear developmental changes at the organ-level.

\vspace{-4mm}
\subsubsection{Dynamic and compositional neural fields.}

Implicit neural representations such as SIREN~\cite{Sitzmann_2020_Neurips,xie2022neural} model high-frequency signals, including geometry, as continuous coordinate-based functions.
Dynamic extensions incorporate deformation fields or velocity-driven ODEs to capture temporal shape evolution~\cite{pumarola2021d,li_2023_neurips,Wang_2025_ICCV,Sun_2023_neurips,Sang_2025_CVPR}.
However, the fine and slender structures common in plants pose challenges for state-of-the-art SDF-based formulations~\cite{luo_leaffit_2026}.
Point-based neural fields~\cite{Prokudin_2023_ICCV,Wu_2024_CVPR} provide higher fidelity for modelling such geometry and have recently been applied to plant dynamics~\cite{luo_grow_2026}.
Additionally, several works have also explored part-aware neural representations for humans and articulated objects.
NASA~\cite{deng_nasa_2020} pioneered this direction using separate neural implicit functions for individual body parts, but such designs scale poorly with increasing numbers of parts and may introduce inconsistencies at part boundaries.
Subsequent approaches, including COAP~\cite{mihajlovic_coap_2022}, DANBO~\cite{su_danbo_2022}, and Neural Body~\cite{peng_neural_2021,peng_implicit_2023}, instead learn shared compositional neural fields conditioned on part-specific embeddings. 
These methods demonstrate the effectiveness of compositional neural fields for modelling articulated motion under \textit{fixed} topology.
In contrast, plant development involves continuous growth and topology changes as new organs emerge.
We extend the compositional neural field paradigm to this setting, conditioning a single shared continuous velocity field on per-organ latent codes to model asynchronous \textit{organ growth} in sparse longitudinal point clouds, capturing the non-rigid, topology-changing nature of plant development.

\section{Method}
\label{sec:method}

\subsection{Intuition}
Plant growth is inherently modular: each organ (stem or leaf) evolves according to its own geometry and structural constraints.
At the same time, organs are not independent; they grow in coordination as parts of a single organism.
Our dynamic plant model is designed to reflect both properties.
We operate on sequences of organ-segmented plant point clouds and align each organ independently into a canonical local coordinate frame.  
This isolates intrinsic per-organ growth patterns from global plant motion and provides stable trajectories for modelling.
We then represent the temporal evolution of all organs with a single shared continuous neural field conditioned on learnable per-organ latent codes.  
The shared network captures geometric and temporal priors common across organs, while the latent codes encode organ identity and growth characteristics.
Together, canonical alignment and latent-conditioned shared dynamics yield a unified yet modular representation of complex, asynchronous plant development.

\subsection{Data preparation and alignment}
\label{subsec:data_preprocessing}

Our input is a time-ordered sequence of 3D point clouds \(\{\mathbf{X}^{(t)}\}_{t=0}^{T}\), each representing a plant at time~\(t\).  
Every point \(\mathbf{x} \in \mathbf{X}^{(t)}\) carries a semantic instance label identifying its organ (stem or leaf).  
These labels provide temporally consistent organ-level trajectories~\cite{li_trackplant3d_2024}.

We represent the plant at time $t$ as a set of organ-specific point clouds \(\mathbf{X}_i^{(t)}\):
\begin{align}
\mathbf{X}^{(t)} = \bigcup_{i=1}^{N^{(t)}} \mathbf{X}_i^{(t)},
\qquad
\mathbf{X}_i^{(t)} \in \mathbb{R}^{N_i^{(t)} \times 3},
\end{align}

where \(N^{(t)}\) is the number of organs observed at time~\(t\) and \(N_i^{(t)}\) is the number of points belonging to organ~\(i\) at that time.

\subsubsection{Leaf-level alignment.}
Leaves exhibit large pose variation due to growth and self-motion.  
To obtain a stable reference frame for intrinsic shape modelling, we apply a PCA-based alignment inspired by physiological leaf coordinates \cite{kirchgessner2003extraktion}. 
Each leaf is oriented such that its principal axis aligns with the global \(Z\)-axis, while its base is translated to the origin.  
This canonicalisation removes global pose variation while preserving intrinsic geometry.
For each timepoint, we store the rigid transformations \(\{\mathbf{R}_i^{(t)}, \mathbf{t}_i^{(t)}\}\) for later reprojection.


\subsection{Organ-conditioned dynamic neural field}
\label{subsec:neural_field}


\subsubsection{Formulation.}
After canonical alignment, organ growth is modelled as a shared continuous neural deformation field conditioned on organ identity.
We define a sinusoidal network~\cite{Sitzmann_2020_Neurips} (SIREN) \(f_{\theta}\) that predicts a time-dependent velocity field:
\begin{align}
f_{\theta}: (\mathbf{x}, t, \mathbf{z}_i) \mapsto \mathbf{v}(\mathbf{x}, t, \mathbf{z}_i) \in \mathbb{R}^3,
\end{align}
where \(\mathbf{x}\) denotes a 3D point in canonical coordinates, \(t \in [0,1]\) denotes normalised time per-organ, and \(\mathbf{z}_i \in \mathbb{R}^{d}\) is a learnable latent code assigned to organ~\(i\).
The latent code is concatenated with spatial and temporal inputs to the network, yielding \([\mathbf{x};\, t;\, \mathbf{z}_i] \in \mathbb{R}^{4+d}\).
The network parameters \(\theta\) are shared across all organs.

\subsubsection{Latent code design.}
We follow the auto-decoder paradigm~\cite{park2019deepsdf}: no encoder is used, and the latent codes \(\{\mathbf{z}_i\}_{i=1}^{N}\) are optimised jointly with the network parameters.
Each organ instance receives its own code, which are initialised from a zero-mean Gaussian \(\mathcal{N}(\mathbf{0}, \sigma_z^2 \mathbf{I})\) and regularised globally toward the origin to prevent unbounded growth:
\begin{align}
\mathcal{L}_{\mathrm{code}} =
\frac{1}{N}\sum_{i=1}^{N} \|\mathbf{z}_i\|_2^2.
\end{align}

\subsubsection{Discrete dynamics.}
Starting from the initial organ point cloud \(\mathbf{X}_i^{(0)}\), we propagate its geometry by accumulating predicted velocities through Euler integration:
\begin{align}
\label{eq:discrete_dynamics}
\mathbf{x}^{(t)} = \mathbf{x}^{(0)}
  + \sum_{\tau = 0,\, \Delta t,\, \dots,\, t-\Delta t}
    f_{\theta}\!\left(\mathbf{x}^{(\tau)}, \tau, \mathbf{z}_i\right) \Delta t,
\end{align}
which approximates the continuous-time formulation:
\begin{align}
\frac{d\mathbf{x}}{dt} = f_{\theta}(\mathbf{x}, t, \mathbf{z}_i)
\quad
\mathbf{x}^{(t)} = \mathbf{x}^{(0)} + \int_{0}^{t} f_{\theta}(\mathbf{x}^{(\tau)}, \tau, \mathbf{z}_i)\, d\tau.
\end{align}

\subsubsection{Training.}
\label{subsubsec:training}
The shared field \(f_{\theta}\) and latent codes \(\{\mathbf{z}_i\}\) are optimised jointly.
At each iteration, we sample a mini-batch of organs \(\mathcal{B} \subset \{1,\dots,N\}\).
For each organ \(i \in \mathcal{B}\), we select source and target timepoints \((s,k)\) with \(k>s\), randomly sample points \(\mathbf{x}^{(s)} \in \mathbf{X}_i^{(s)}\), and propagate them forward through the learned velocity field.

\subsubsection{Chamfer supervision.}
We supervise the predicted trajectory using bidirectional Chamfer distance
computed at every predicted timestep:
\begin{align}
\mathcal{L}_{\mathrm{CD}}^{(i)} 
=
\frac{1}{k-s}
\sum_{t=s+1}^{k}
\mathrm{CD}\!\left(
\hat{\mathbf{X}}_i^{(t)},
\mathbf{X}_i^{(t)}
\right).
\end{align}

The Chamfer distance between two point sets \(\hat{\mathbf{X}}\) and
\(\mathbf{X}\) is defined as
\begin{align}
\mathrm{CD}(\hat{\mathbf{X}},\mathbf{X}) =
\frac{1}{|\hat{\mathbf{X}}|}
\sum_{\mathbf{x} \in \hat{\mathbf{X}}}
\min_{\mathbf{y} \in \mathbf{X}}
\|\mathbf{x}-\mathbf{y}\|_2^2
+
\frac{1}{|\mathbf{X}|}
\sum_{\mathbf{y} \in \mathbf{X}}
\min_{\mathbf{x} \in \hat{\mathbf{X}}}
\|\mathbf{x}-\mathbf{y}\|_2^2.
\end{align}

\subsubsection{Total objective.}
The Chamfer loss is computed per organ and averaged
over the sampled mini-batch, yielding the following final training objective:
\begin{align}
\mathcal{L}_{\mathrm{total}}
=
\frac{1}{|\mathcal{B}|}
\sum_{i \in \mathcal{B}}
\mathcal{L}_{\mathrm{CD}}^{(i)}
+
\lambda_{\mathrm{code}} \mathcal{L}_{\mathrm{code}} .
\end{align}



\subsection{Full-plant dynamics}
\label{subsec:full_plant_dynamics}


\subsubsection{Composition.}
A complete 4D reconstruction of the plant requires merging the continuous organ trajectories into a single, temporally coherent structure.
For each organ \(i\), we select its earliest observed frame and use the corresponding point cloud \(\mathbf{X}_i^{(0)}\) as initialisation.
Note that our model does not \emph{predict} organ emergence; instead, each newly observed organ is incorporated into the shared dynamical representation upon its first observation.
This sidesteps the ill-posed problem of synthesizing unseen geometry while still yielding a coherent, topology-changing full-plant reconstruction.
Given the shared neural field \(f_{\theta}\) and the organ's latent code \(\mathbf{z}_i\), we propagate the geometry forward through the field, integrating its velocity over time:
\begin{align}
\hat{\mathbf{X}}_i^{(t+\Delta t)} = \hat{\mathbf{X}}_i^{(t)} + f_{\theta}(\hat{\mathbf{X}}_i^{(t)}, t, \mathbf{z}_i)\,\Delta t,
\end{align}
producing a continuous temporal trajectory within the canonical coordinate frame.  
Each predicted organ state \(\hat{\mathbf{X}}_i^{(t)}\) is then transformed back to the global plant space via its rigid alignment parameters:
\begin{align}
\hat{\mathbf{X}}_{i,\text{global}}^{(t)} = \hat{\mathbf{X}}_i^{(t)} \mathbf{R}_i^{(t)\top} + \mathbf{t}_i^{(t)}.
\end{align}
The rigid poses \(\{\mathbf{R}_i^{(t)}, \mathbf{t}_i^{(t)}\}\) are recovered from the canonical alignment of each observed scan (\autoref{subsec:data_preprocessing}); for frames with no observed scan, such as dense interpolation, they are obtained via SLERP on rotations and linear interpolation on translations.

\subsubsection{Plant synthesis.}
The full plant model at time \(t\) is obtained by combining all reconstructed organ point sets:
\begin{align}
\hat{\mathbf{X}}^{(t)} = \bigcup_{i=1}^{N^{(t)}} \hat{\mathbf{X}}_{i,\text{global}}^{(t)}.
\end{align}
This synthesis yields a complete 4D plant trajectory that integrates asynchronous, organ-specific growth patterns into a coherent dynamic representation.  

\section{Experiments}
\label{sec:experiments}

\subsection{Experimental setup}

\subsubsection{Dataset and evaluation protocol.}
We evaluate on the TrackPlant3D dataset \cite{li_trackplant3d_2024}, comprising daily LiDAR scans of four species (maize, sorghum, tobacco, tomato) with per-organ instance labels, collected from two public sources~\cite{schunck_pheno4d_2021, Conn_2017_lidar}.
We use eight test and four validation sequences (one per species, for hyperparameter tuning), spanning 5--17 timepoints each; to enable tracking evaluation, we additionally annotated leaf tips on all scans where they could be clearly identified.
We report Chamfer distance (CD, mm$^2$) for geometric fit and End-Point Error (EPE, mm) on the annotated leaf tips for tracking accuracy, averaged over timesteps per sequence and computed on test points only.
Unless otherwise stated, all reported metrics are averaged over five independent runs, and predicted geometry is reprojected using the available ground-truth per-frame poses so that the metrics isolate shape and growth from pose.
Further dataset, implementation, and baseline details are in the supplementary material.

Following prior neural-field works~\cite{Zhang2024_DOMA_CVPR}, we randomly split each organ's point cloud into 50\% train and 50\% test points independently at each timestep, ensuring that the 234 leaf-tip points are always in the test set.
This split is fixed and shared across all methods.
At inference we predict \emph{full trajectories}: from the first observed frame ($t_0$) we integrate the deformation field multi-step and evaluate at every timestep against the ground truth, rather than only single-step-ahead predictions.

\subsubsection{Baselines.}
We compare against general 4D shape-interpolation methods that operate on the \emph{full plant} without segmentation input: CanFields~\cite{Wang_2025_ICCV}, Neural Velocity Fields (NVFi)~\cite{li_2023_neurips}, Neural Diffeomorphic Flow (NDF)~\cite{Sun_2022_CVPR}, a 4D SDF model (DSR)~\cite{Sun_2023_neurips}, and chained Dynamic Point Fields (DPF)~\cite{Prokudin_2023_ICCV}, where each frame transition is modelled by a separate MLP (also evaluated per organ).
As \emph{part-aware} baselines that use the same ground-truth organ segmentation as our method, we extend the compositional COAP~\cite{mihajlovic_coap_2022} encoder-decoder into a dynamic deformation field trained like GrowFields.

\begin{table}[tb]
\centering
\caption{\textbf{Full plant reconstruction results.}
Mean per-sequence temporal metrics across eight growth sequences: CD (mm$^2$) for geometric accuracy and EPE (mm) on annotated leaf tips for tracking accuracy.
Full-plant baselines take no segmentation input, whereas the part-aware COAP and our models use the same ground-truth organ segmentation.
Our full model achieves the best overall performance; DSR is a 4D-implicit method with no deformation field, so EPE is undefined.
Colours indicate column-wise ranking: \colorbox[RGB]{\colorfirst}{1st}, \colorbox[RGB]{\colorsecond}{2nd}, and \colorbox[RGB]{\colorthird}{3rd}. All metrics are averaged over five independent runs.
}
\label{tab:results_per_plant}
\resizebox{\textwidth}{!}{
\begin{tabular}{l cc cc cc cc cc cc cc cc cc}
\toprule
 & \multicolumn{2}{c}{MaC2} & \multicolumn{2}{c}{MaC3} & \multicolumn{2}{c}{SoC2} & \multicolumn{2}{c}{SoH2} & \multicolumn{2}{c}{TbC1} & \multicolumn{2}{c}{TbS3} & \multicolumn{2}{c}{Tm1H3} & \multicolumn{2}{c}{Tm1S1} & \multicolumn{2}{c}{\textit{Mean}} \\
\cmidrule(lr){2-3} \cmidrule(lr){4-5} \cmidrule(lr){6-7} \cmidrule(lr){8-9} \cmidrule(lr){10-11} \cmidrule(lr){12-13} \cmidrule(lr){14-15} \cmidrule(lr){16-17} \cmidrule(lr){18-19}
Method & CD $\downarrow$ & EPE $\downarrow$ & CD $\downarrow$ & EPE $\downarrow$ & CD $\downarrow$ & EPE $\downarrow$ & CD $\downarrow$ & EPE $\downarrow$ & CD $\downarrow$ & EPE $\downarrow$ & CD $\downarrow$ & EPE $\downarrow$ & CD $\downarrow$ & EPE $\downarrow$ & CD $\downarrow$ & EPE $\downarrow$ & CD $\downarrow$ & EPE $\downarrow$ \\
\midrule
\multicolumn{19}{l}{\textit{Full-plant baselines}} \\
NDF \cite{Sun_2022_CVPR} & 14,496.5 & 73.6 & 1,227.9 & 95.8 & 1,313.4 & 54.9 & 303.0 & 54.3 & 290.2 & 14.0 & 135.6 & 17.9 & 14.1 & 10.4 & 52.7 & 22.0 & 2,229.2 & 42.9 \\
NVFi \cite{li_2023_neurips} & 2,178.9 & 72.2 & 1,495.3 & 97.4 & 873.0 & 56.3 & 161.8 & 56.0 & 103.0 & 15.1 & 86.0 & 18.2 & 32.6 & 9.4 & 137.3 & 21.9 & 633.5 & 43.3 \\
DSR \cite{Sun_2023_neurips} & 481.3 & --- & 194.9 & --- & 186.4 & --- & 61.1 & --- & 29.9 & --- & 25.0 & --- & 13.1 & --- & 84.0 & --- & 134.5 & --- \\
CanFields \cite{Wang_2025_ICCV} & 2,025.7 & 77.6 & 1,774.8 & 98.9 & 611.4 & 56.4 & 267.6 & 59.5 & 102.1 & 16.6 & 67.3 & 21.1 & 23.0 & 10.4 & 53.5 & 14.4 & 615.7 & 44.4 \\
DPF \cite{Prokudin_2023_ICCV} &  8.17 & 23.18 & 5.00 & 5.30 & \cellcolor{third} 1.21 & \cellcolor{third} 5.78 & \cellcolor{second} 0.92 & 57.22 & \cellcolor{best} 0.36 & 7.58 & \cellcolor{second} 0.46 & 16.21 & \cellcolor{third} 0.17 & 1.93 & 3.47 & 3.97 & 2.47 & 15.15 \\ 
\midrule
\multicolumn{19}{l}{\textit{Part-aware methods}} \\
COAP \cite{mihajlovic_coap_2022} & \cellcolor{third} 4.84 & \cellcolor{third} 7.57 & \cellcolor{third} 2.48 & \cellcolor{third} 3.87 & 1.31 & 7.76 & 1.02 & \cellcolor{third} 5.61 & 1.03 & \cellcolor{second} 3.22 & 0.65 & \cellcolor{second} 2.55 & 0.38 & \cellcolor{third} 1.40 & \cellcolor{third} 1.82 & \cellcolor{third} 1.88 & \cellcolor{third} 1.69 & \cellcolor{third} 4.23 \\
Ours (per-organ) & \cellcolor{best} 2.41 & \cellcolor{second} 3.68 & \cellcolor{second} 1.52 & \cellcolor{second} 2.07 & \cellcolor{second} 0.87 & \cellcolor{second} 1.47 & \cellcolor{third} 1.01 & \cellcolor{second} 3.44 & \cellcolor{third} 0.70 & \cellcolor{third} 3.28 & \cellcolor{third} 0.48 & \cellcolor{third} 2.67 & \cellcolor{second} 0.13 & \cellcolor{best} 0.74 & \cellcolor{second} 0.33 & \cellcolor{second} 1.59 & \cellcolor{second} 0.93 & \cellcolor{second} 2.37 \\
\textbf{Ours (full)} & \cellcolor{second} 2.51 & \cellcolor{best} 2.38 & \cellcolor{best} 1.43 & \cellcolor{best} 2.01 & \cellcolor{best} 0.69 & \cellcolor{best} 1.34 & \cellcolor{best} 0.74 & \cellcolor{best} 0.97 & \cellcolor{second} 0.59 & \cellcolor{best} 1.66 & \cellcolor{best} 0.45 & \cellcolor{best} 1.30 & \cellcolor{best} 0.12 & \cellcolor{second} 0.80 & \cellcolor{best} 0.32 & \cellcolor{best} 1.34 & \cellcolor{best} 0.86 & \cellcolor{best} 1.47 \\

\bottomrule
\end{tabular}
}
\end{table}

\subsection{Results}

\begin{figure}[t]
    \centering
    \begin{subfigure}[t]{0.49\textwidth}
      \includegraphics[width=\textwidth]{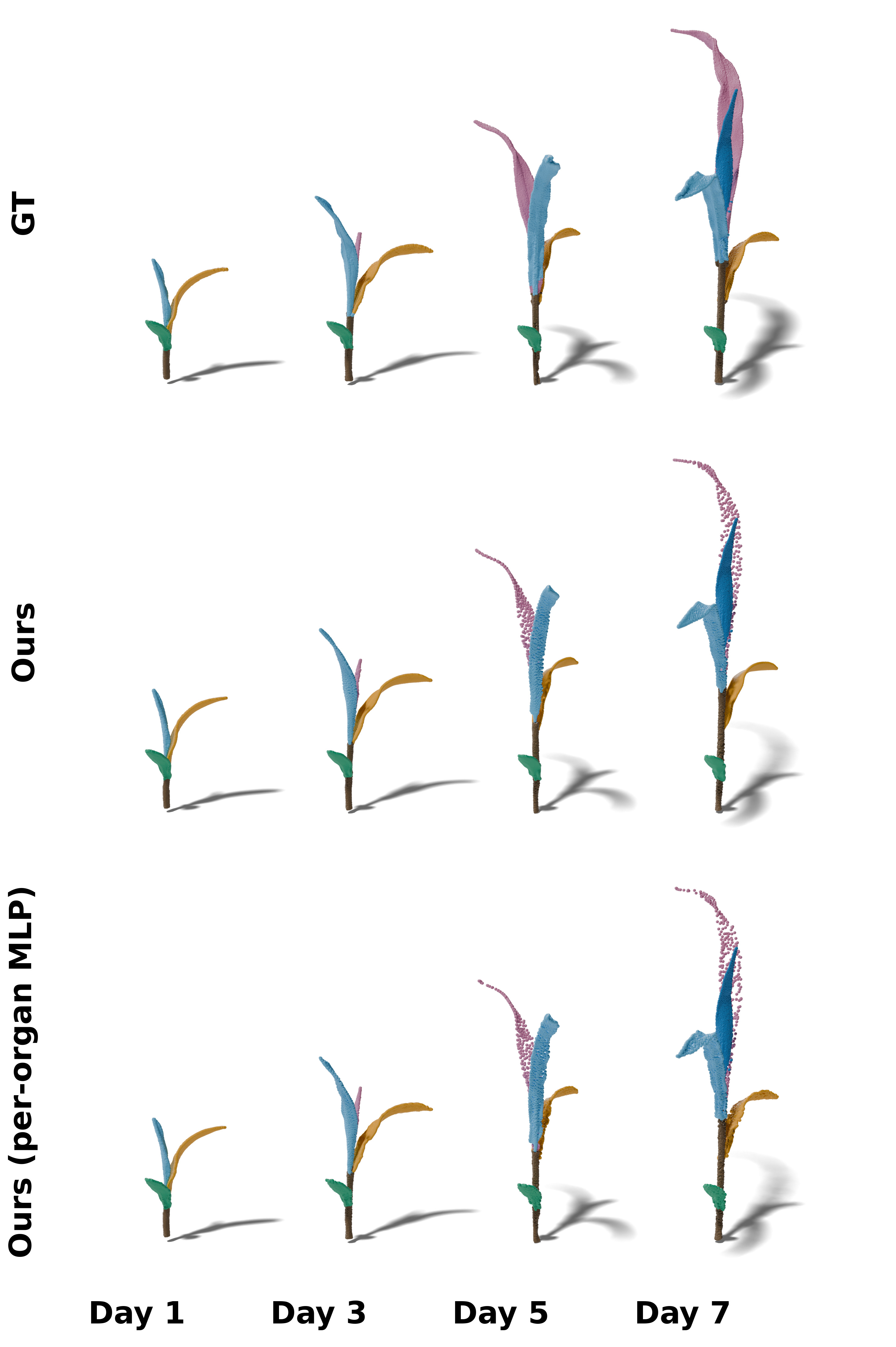}
    \end{subfigure}
    \hfill
    \begin{subfigure}[t]{0.49\textwidth}
      \includegraphics[width=\textwidth]{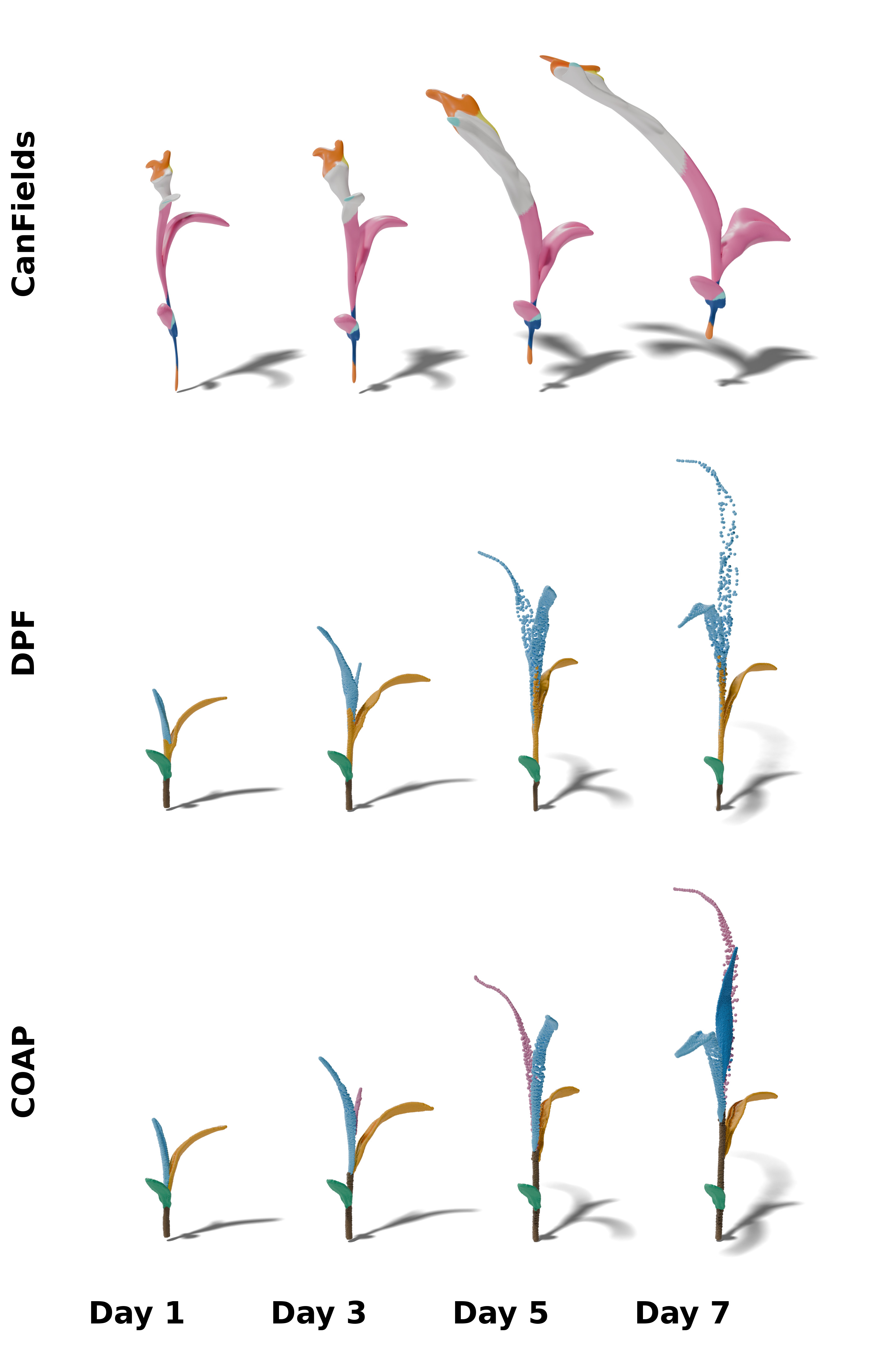}
    \end{subfigure}
    \caption{\textbf{Qualitative full-plant reconstruction.} Colours denote organ identities from the ground-truth segmentation, except for CanFields~\cite{Wang_2025_ICCV}, where they denote the parts discovered by its own internal motion-segmentation. Compared to per-organ MLPs, our shared latent-conditioned field resolves part boundaries correctly, illustrating the effectiveness of the latent-code formulation.}
   \label{fig:fullplant_methods}
  \end{figure}

\textbf{Plant-level results.}
We present full plant reconstruction results in \autoref{tab:results_per_plant} and qualitative comparisons in \autoref{fig:fullplant_methods}.
Methods that model scene dynamics with global deformation or motion fields, such as NDF~\cite{Sun_2022_CVPR} and NVFi~\cite{li_2023_neurips}, struggle with the rapidly changing topology of plant growth. 
When new organs emerge, these models must explain previously unseen geometry through a continuous deformation of existing structure, which often leads to unstable or degenerate solutions.

SDF-based approaches such as DSR~\cite{Sun_2023_neurips} and canonical-shape interpolation methods such as CanFields~\cite{Wang_2025_ICCV} improve geometric fitting in some cases but still produce large tracking errors. 
These methods favour coherent global surface evolution and therefore struggle to model thin structures such as leaves and stems whose motion and growth are largely organ-specific.

Point-based models provide a stronger baseline. Dynamic Point Fields (DPF) ~\cite{Prokudin_2023_ICCV} represent geometry with explicit point primitives while learning an implicit deformation model, which improves reconstruction accuracy. 
However, without an explicit mechanism for modelling organ emergence and independent growth, performance still degrades when plant structure changes substantially across timesteps.

Compositional models are more robust in this setting. Our point-based adaptation of the part-aware COAP framework~\cite{mihajlovic_coap_2022} retains the part decomposition but replaces the occupancy representation with per-part point deformation fields. 
This decomposition better reflects plant morphology and allows individual organs to evolve independently.

Our method further improves on these approaches by combining per-organ canonicalisation with a shared latent-conditioned deformation field.
This formulation models organ-specific motion while capturing global correlations between plant parts. 
As shown in \autoref{tab:results_per_plant}, the full model achieves the best overall performance across sequences and metrics, improving both reconstruction accuracy (CD) and leaf-tip tracking accuracy (EPE).

\begin{table}[tb]
\centering
\caption{\textbf{Per-organ reconstruction results.}
We report mean temporal metrics for individual organ sequences grouped by their parent plant sequence.
The \textit{Mean} column reports the average across the eight plant sequences (each plant weighted equally).
Compared to full-plant evaluation (Table~1), most baselines show only modest improvements despite operating on isolated organs. Our full model achieves the best overall performance, with the lowest mean CD (0.87) and EPE (1.62), indicating improved geometric reconstruction and leaf-tip tracking across organ sequences. Note that DSR is a 4D-implicit method with no deformation field, so point tracking (EPE) is undefined.
All metrics are averaged over five independent runs.
}
\label{tab:results_per_organ}
\resizebox{\textwidth}{!}{%
\begin{tabular}{l cc cc cc cc cc cc cc cc cc}
\toprule
 & \multicolumn{2}{c}{MaC2} & \multicolumn{2}{c}{MaC3} & \multicolumn{2}{c}{SoC2} & \multicolumn{2}{c}{SoH2} & \multicolumn{2}{c}{TbC1} & \multicolumn{2}{c}{TbS3} & \multicolumn{2}{c}{Tm1H3} & \multicolumn{2}{c}{Tm1S1} & \multicolumn{2}{c}{\textit{Mean}} \\
\cmidrule(lr){2-3} \cmidrule(lr){4-5} \cmidrule(lr){6-7} \cmidrule(lr){8-9} \cmidrule(lr){10-11} \cmidrule(lr){12-13} \cmidrule(lr){14-15} \cmidrule(lr){16-17} \cmidrule(lr){18-19}
Method & CD $\downarrow$ & EPE $\downarrow$ & CD $\downarrow$ & EPE $\downarrow$ & CD $\downarrow$ & EPE $\downarrow$ & CD $\downarrow$ & EPE $\downarrow$ & CD $\downarrow$ & EPE $\downarrow$ & CD $\downarrow$ & EPE $\downarrow$ & CD $\downarrow$ & EPE $\downarrow$ & CD $\downarrow$ & EPE $\downarrow$ & CD $\downarrow$ & EPE $\downarrow$ \\
\midrule
NDF \cite{Sun_2022_CVPR}  & 484.6 & 65.9 & 317.0 & 104.4 & 45.9 & 69.7 & 17.6 & 29.8 & 29.5 & 14.0 & 15.2 & 6.8 & 1.5 & 2.7 & 3.6 & 6.8 & 114.4 & 37.5 \\
NVFi \cite{li_2023_neurips} & 540.1 & 68.7 & 190.1 & 100.4 & 37.5 & 75.0 & 13.7 & 28.6 & 13.9 & 15.3 & 7.2 & 6.4 & 1.8 & 2.5 & 5.3 & 7.7 & 101.2 & 38.1 \\
DSR \cite{Sun_2023_neurips} & 41.9 & --- & 21.6 & --- & 26.9 & --- & 10.1 & --- & 15.3 & --- & 6.5 & --- & 1.3 & --- & 2.1 & --- & 15.7 & --- \\
CanFields \cite{Wang_2025_ICCV} & 245.0 & 52.1 & 355.7 & 77.1 & 174.8 & 55.3 & 294.6 & 31.1 & 18.9 & 7.1 & 6.6 & 4.7 & 2.0 & 2.7 & 6.1 & 5.1 & 138.0 & 29.4 \\
DPF \cite{Prokudin_2023_ICCV} & \cellcolor{third} 2.32 & \cellcolor{best} 2.05 & \cellcolor{third} 1.67 & \cellcolor{second} 2.60 & \cellcolor{third} 1.30 & \cellcolor{best} 1.43 & \cellcolor{second} 0.87 & \cellcolor{second} 1.43 & \cellcolor{third} 0.69 & \cellcolor{second} 3.18 & \cellcolor{third} 0.56 & \cellcolor{second} 1.57 & \cellcolor{third} 0.30 & 2.01 & \cellcolor{third} 0.68 & 2.23 & \cellcolor{third} 1.05 & \cellcolor{second} 2.06 \\
COAP \cite{mihajlovic_coap_2022} & 4.17 & 8.20 & 2.55 & 5.68 & 2.22 & 16.63 & \cellcolor{third} 0.93 & 8.82 & 0.99 & 5.01 & 0.64 & 2.02 & 0.42 & \cellcolor{third} 1.75 & 1.60 & \cellcolor{third} 1.93 & 1.69 & 6.26 \\
Ours (per-organ MLPs) & \cellcolor{best} 2.10 & \cellcolor{third} 3.47 & \cellcolor{second} 1.52 & \cellcolor{third} 2.89 & \cellcolor{second} 1.13 & \cellcolor{third} 1.59 & 0.96 & \cellcolor{third} 2.34 & \cellcolor{second} 0.56 & \cellcolor{third} 4.13 & \cellcolor{second} 0.50 & \cellcolor{third} 1.88 & \cellcolor{second} 0.24 & \cellcolor{second} 1.42 & \cellcolor{second} 0.51 & \cellcolor{second} 1.84 & \cellcolor{second} 0.94 & \cellcolor{third} 2.44 \\
\textbf{Ours (full)} & \cellcolor{second} 2.26 & \cellcolor{second} 2.59 & \cellcolor{best} 1.48 & \cellcolor{best} 2.46 & \cellcolor{best} 0.93 & \cellcolor{second} 1.51 & \cellcolor{best} 0.64 & \cellcolor{best} 0.85 & \cellcolor{best} 0.49 & \cellcolor{best} 2.04 & \cellcolor{best} 0.45 & \cellcolor{best} 1.20 & \cellcolor{best} 0.20 & \cellcolor{best} 0.98 & \cellcolor{best} 0.46 & \cellcolor{best} 1.33 & \cellcolor{best} 0.87 & \cellcolor{best} 1.62 \\
\bottomrule
\end{tabular}
}
\end{table}

\textbf{Organ-level results.}
Since several baselines struggle in the full-plant setting (\autoref{tab:results_per_plant}), we additionally test the models by providing per-organ sequences independently as input, in order to probe their behaviour in a reduced, simpler, topology-invariant setting.
\autoref{tab:results_per_organ} reports reconstruction metrics computed per organ sequence.
To ensure fair comparison with baselines, we consider only organ sequences with at least three frames.
This reduces the evaluation set from 49 to 45 sequences and from a total of 192 to 189 leaf-tip error measurements.

Despite operating on isolated organs, most baselines improve only modestly compared to the full-plant setting. 
For example, DPF achieves a mean CD of 1.05 and EPE of 2.06, while COAP obtains 1.69 CD and 6.26 EPE. 
Our per-organ MLP variant improves these results to 0.94 CD and 2.44 EPE.
The full model achieves the best overall performance with a mean CD of 0.87 and EPE of 1.62. 
This indicates that modelling interactions between organs remains beneficial even when evaluation is performed at the organ level.

\textbf{Interpolation.}
To evaluate temporal generalisation, we train the models using only every second timestep and evaluate reconstruction on both observed and interpolated frames.
Results are summarised in \autoref{tab:interp_metrics} and \autoref{tab:interp_metrics_perorgan}, where we group metrics separately for \emph{observed} (training) frames and the held-out \emph{interpolated} frames.

Because most sequences are short, typically providing only four to five training timesteps, interpolation is challenging. 
Under these conditions, DPF (per-organ) remains a strong baseline and achieves the best overall interpolation performance (mean CD 10.24, EPE 4.00). This behaviour reflects the strong local fitting ability of DPF, whereas our model prioritises globally consistent growth dynamics. Our model remains competitive and performs particularly well on longer sequences such as Tm1H3, which contains 16 timesteps (8 used for training). 
These results indicate that the proposed representation captures smooth temporal dynamics while maintaining accurate geometric reconstruction.
Qualitative interpolation and point-tracking comparisons are provided in the supplementary material.

\begin{table}[tb]
\centering
\caption{\textbf{Full plant interpolation results.}
Models are trained using every second timestep and evaluated on both observed (training) and interpolated frames.
DPF (per-organ) achieves the strongest interpolation performance overall, while our full model remains competitive among continuous neural-field approaches and performs particularly well on longer sequences.
All metrics are averaged over five independent runs.
}
\label{tab:interp_metrics}
\resizebox{\textwidth}{!}{
\begin{tabular}{l cc cc cc cc cc cc cc cc cc}
\toprule
 & \multicolumn{2}{c}{MaC2} & \multicolumn{2}{c}{MaC3} & \multicolumn{2}{c}{SoC2} & \multicolumn{2}{c}{SoH2} & \multicolumn{2}{c}{TbC1} & \multicolumn{2}{c}{TbS3} & \multicolumn{2}{c}{Tm1H3} & \multicolumn{2}{c}{Tm1S1} & \multicolumn{2}{c}{\textit{Mean}} \\
\cmidrule(lr){2-3} \cmidrule(lr){4-5} \cmidrule(lr){6-7} \cmidrule(lr){8-9} \cmidrule(lr){10-11} \cmidrule(lr){12-13} \cmidrule(lr){14-15} \cmidrule(lr){16-17} \cmidrule(lr){18-19}
Method & CD $\downarrow$ & EPE $\downarrow$ & CD $\downarrow$ & EPE $\downarrow$ & CD $\downarrow$ & EPE $\downarrow$ & CD $\downarrow$ & EPE $\downarrow$ & CD $\downarrow$ & EPE $\downarrow$ & CD $\downarrow$ & EPE $\downarrow$ & CD $\downarrow$ & EPE $\downarrow$ & CD $\downarrow$ & EPE $\downarrow$ & CD $\downarrow$ & EPE $\downarrow$ \\
\midrule
\multicolumn{19}{c}{\textit{Training frames}} \\
COAP~\cite{mihajlovic_coap_2022} & 3.20 & 11.87 & \cellcolor{third} 0.82 & \cellcolor{third} 3.08 & 3.04 & 9.60 & 1.11 & \cellcolor{second} 2.20 & 0.91 & \cellcolor{second} 2.39 & \cellcolor{third} 0.56 & 2.34 & 0.15 & 2.46 & 0.59 & \cellcolor{second} 1.40 & 1.30 & 4.42 \\
DPF (per-organ) & \cellcolor{third} 2.62 & \cellcolor{best} 1.94 & \cellcolor{second} 0.77 & \cellcolor{best} 1.67 & \cellcolor{second} 0.51 & \cellcolor{second} 0.85 & \cellcolor{second} 0.63 & \cellcolor{third} 2.87 & \cellcolor{third} 0.83 & \cellcolor{best} 1.98 & 0.57 & \cellcolor{best} 1.58 & \cellcolor{third} 0.13 & \cellcolor{third} 1.26 & \cellcolor{third} 0.41 & \cellcolor{third} 1.67 & \cellcolor{second} 0.81 & \cellcolor{best} 1.73 \\
Ours (per-organ MLPs) & \cellcolor{second} 2.59 & \cellcolor{third} 2.21 & 0.97 & 5.49 & \cellcolor{third} 0.59 & \cellcolor{third} 1.28 & \cellcolor{third} 0.72 & 4.73 & \cellcolor{second} 0.77 & \cellcolor{third} 3.44 & \cellcolor{second} 0.47 & \cellcolor{third} 2.05 & \cellcolor{second} 0.12 & \cellcolor{second} 1.14 & \cellcolor{second} 0.37 & 1.93 & \cellcolor{third} 0.82 & \cellcolor{third} 2.78 \\
\textbf{Ours (full)} & \cellcolor{best} 2.24 & \cellcolor{second} 2.14 & \cellcolor{best} 0.70 & \cellcolor{second} 1.72 & \cellcolor{best} 0.42 & \cellcolor{best} 0.71 & \cellcolor{best} 0.44 & \cellcolor{best} 0.56 & \cellcolor{best} 0.57 & 5.39 & \cellcolor{best} 0.39 & \cellcolor{second} 1.70 & \cellcolor{best} 0.10 & \cellcolor{best} 0.58 & \cellcolor{best} 0.35 & \cellcolor{best} 1.24 & \cellcolor{best} 0.65 & \cellcolor{second} 1.75 \\
\midrule
\multicolumn{19}{c}{\textit{Interpolated frames}} \\
COAP~\cite{mihajlovic_coap_2022} & 19.30 & 14.63 & 9.57 & \cellcolor{third} 12.97 & 25.42 & 10.20 & 31.07 & 9.97 & 4.16 & \cellcolor{second} 3.90 & \cellcolor{third} 1.88 & 2.82 & \cellcolor{best} 0.83 & 2.55 & 1.62 & \cellcolor{second} 1.63 & 11.73 & 7.33 \\
DPF (per-organ) & \cellcolor{best} 17.75 & \cellcolor{third} 8.06 & \cellcolor{best} 8.93 & \cellcolor{best} 4.72 & \cellcolor{best} 18.38 & \cellcolor{best} 5.26 & \cellcolor{second} 29.38 & \cellcolor{best} 5.74 & \cellcolor{third} 3.23 & \cellcolor{best} 3.13 & \cellcolor{third} 1.88 & \cellcolor{best} 1.89 & \cellcolor{second} 0.89 & \cellcolor{third} 1.63 & \cellcolor{second} 1.47 & \cellcolor{best} 1.60 & \cellcolor{best} 10.24 & \cellcolor{best} 4.00 \\
Ours (per-organ MLPs) & \cellcolor{second} 18.00 & \cellcolor{second} 8.04 & \cellcolor{third} 9.47 & 13.32 & \cellcolor{second} 19.17 & \cellcolor{third} 5.55 & \cellcolor{third} 29.42 & \cellcolor{third} 9.73 & \cellcolor{best} 2.84 & \cellcolor{third} 4.59 & \cellcolor{second} 1.76 & \cellcolor{third} 2.57 & \cellcolor{third} 0.91 & \cellcolor{second} 1.56 & \cellcolor{third} 1.53 & 2.28 & \cellcolor{third} 10.39 & \cellcolor{third} 5.96 \\
\textbf{Ours (full)} & \cellcolor{third} 18.01 & \cellcolor{best} 7.76 & \cellcolor{second} 9.18 & \cellcolor{second} 11.82 & \cellcolor{third} 19.53 & \cellcolor{second} 5.34 & \cellcolor{best} 28.60 & \cellcolor{second} 9.69 & \cellcolor{second} 2.93 & 6.49 & \cellcolor{best} 1.75 & \cellcolor{second} 2.48 & \cellcolor{second} 0.89 & \cellcolor{best} 1.12 & \cellcolor{best} 1.42 & \cellcolor{third} 1.88 & \cellcolor{second} 10.29 & \cellcolor{second} 5.82 \\
\bottomrule
\end{tabular}
}
\end{table}

\begin{figure}[t]
\centering
\begin{tikzpicture}
  \node[anchor=south west, inner sep=0] (img) {\includegraphics[width=1.2\linewidth]{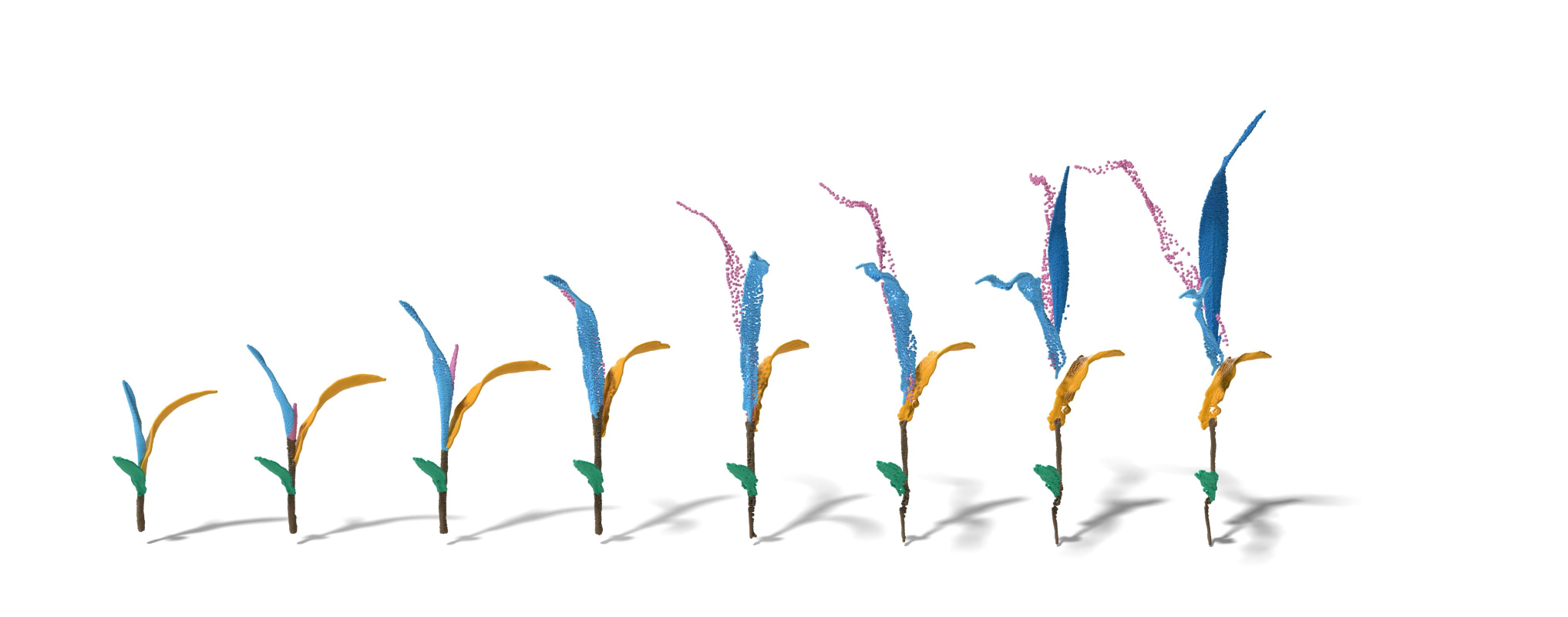}};
  \begin{scope}[x={(img.south east)},y={(img.north west)}]
    \draw[eccvblue, very thick, dashed, rounded corners=3pt] (0.52,0.03) rectangle (0.85,0.97);
    \node[eccvblue, anchor=north west, font=\small\bfseries] at (0.535,0.94) {Extrapolation};
  \end{scope}
\end{tikzpicture}
\caption{\textbf{Qualitative extrapolation.} GrowFields prediction on \texttt{maize\_control\_plant2}, trained on all but the last three timesteps and rolled out from $t_0$ across the full sequence (left to right); colours denote organ identities, and the dashed box marks the three held-out frames evaluated for extrapolation. Although the model predicts plausible continued organ elongation into the held-out frames; without generative priors, geometric accuracy rapidly degrades (see \autoref{sec:limitations}).}
\label{fig:extrapolation_qual}
\end{figure}

\textbf{Extrapolation.}
We additionally test longer-horizon extrapolation by training on all timesteps \emph{except} the last three and evaluating exclusively on those held-out final frames (SoH2 is excluded, as it has only five timesteps).
Predictions are multi-step trajectories rolled out from $t_0$ beyond the training horizon.
As shown in \autoref{tab:extrapolation}, GrowFields outperforms COAP across all sequences; the large absolute errors on some sequences are partly explained by new organs emerging within the held-out frames, a limitation we discuss in \autoref{sec:limitations}. A qualitative example is shown in \autoref{fig:extrapolation_qual}.

\begin{table}[htbp]
\centering
\caption{\textbf{Extrapolation results.} Models are trained on all but the last three timesteps and evaluated only on those held-out final frames; predictions are multi-step rollouts from $t_0$. We report CD~$\downarrow$ / EPE~$\downarrow$. GrowFields outperforms COAP on every sequence. All metrics are averaged over five independent runs.}
\label{tab:extrapolation}
\resizebox{\textwidth}{!}{
\begin{tabular}{l cc cc cc cc cc cc cc cc}
\toprule
 & \multicolumn{2}{c}{MaC2} & \multicolumn{2}{c}{MaC3} & \multicolumn{2}{c}{SoC2} & \multicolumn{2}{c}{TbC1} & \multicolumn{2}{c}{TbS3} & \multicolumn{2}{c}{Tm1H3} & \multicolumn{2}{c}{Tm1S1} & \multicolumn{2}{c}{\textit{Mean}} \\
\cmidrule(lr){2-3} \cmidrule(lr){4-5} \cmidrule(lr){6-7} \cmidrule(lr){8-9} \cmidrule(lr){10-11} \cmidrule(lr){12-13} \cmidrule(lr){14-15} \cmidrule(lr){16-17}
Method & CD $\downarrow$ & EPE $\downarrow$ & CD $\downarrow$ & EPE $\downarrow$ & CD $\downarrow$ & EPE $\downarrow$ & CD $\downarrow$ & EPE $\downarrow$ & CD $\downarrow$ & EPE $\downarrow$ & CD $\downarrow$ & EPE $\downarrow$ & CD $\downarrow$ & EPE $\downarrow$ & CD $\downarrow$ & EPE $\downarrow$ \\
\midrule
\multicolumn{17}{c}{\textit{Training frames}} \\
COAP~\cite{mihajlovic_coap_2022} & 1.34 & 12.26 & 0.57 & 5.01 & 0.49 & \cellcolor{best} 1.02 & 0.18 & 1.13 & 0.26 & \cellcolor{best} 1.24 & 0.22 & 1.15 & 0.54 & 1.54 & 0.51 & 3.34 \\
\textbf{Ours} & \cellcolor{best} 1.06 & \cellcolor{best} 1.81 & \cellcolor{best} 0.40 & \cellcolor{best} 0.92 & \cellcolor{best} 0.39 & 1.08 & \cellcolor{best} 0.15 & \cellcolor{best} 0.80 & \cellcolor{best} 0.18 & 1.67 & \cellcolor{best} 0.14 & \cellcolor{best} 0.79 & \cellcolor{best} 0.30 & \cellcolor{best} 1.33 & \cellcolor{best} 0.37 & \cellcolor{best} 1.20 \\
\midrule
\multicolumn{17}{c}{\textit{Extrapolated frames}} \\
COAP~\cite{mihajlovic_coap_2022} & 768.08 & 47.17 & 1632.07 & 103.33 & 620.43 & 42.77 & 79.27 & 9.23 & 15.82 & 5.04 & 8.59 & 5.92 & 33.40 & 6.25 & 451.09 & 31.39 \\
\textbf{Ours} & \cellcolor{best} 356.13 & \cellcolor{best} 20.96 & \cellcolor{best} 321.08 & \cellcolor{best} 69.77 & \cellcolor{best} 337.39 & \cellcolor{best} 27.04 & \cellcolor{best} 15.38 & \cellcolor{best} 4.06 & \cellcolor{best} 8.09 & \cellcolor{best} 4.14 & \cellcolor{best} 2.80 & \cellcolor{best} 2.52 & \cellcolor{best} 6.56 & \cellcolor{best} 4.46 & \cellcolor{best} 149.63 & \cellcolor{best} 18.99 \\
\bottomrule
\end{tabular}
}
\end{table}

\begin{figure}[t]
\centering
\includegraphics[width=1.2\linewidth, trim={0 35 0 24}, clip]{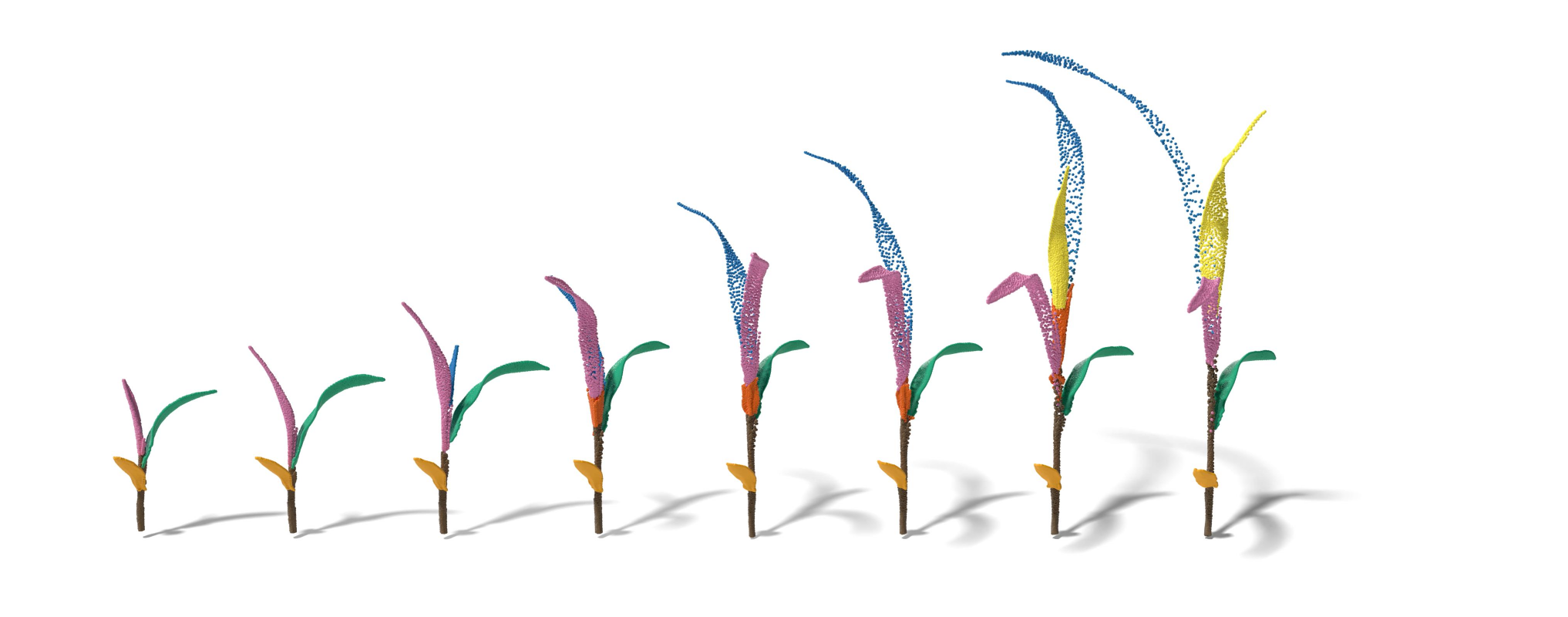}
\caption{\textbf{Qualitative result under automated segmentation.} GrowFields prediction on \texttt{maize\_control\_plant2} using \emph{automatically} segmented organ labels as input; colours denote the input organ identities. Despite noisier labels than the ground-truth segmentation, the method still recovers coherent organ-level growth.}
\label{fig:autoseg_qual}
\end{figure}

\subsection{Analysis}

\textbf{Ablation study.}
We analyse the contribution of the main design components in \autoref{tab:ablations}.
Removing the organ's latent code leads to the largest drop in reconstruction accuracy, increasing the mean error from 0.86/1.47 to 1.05/3.44 (CD/EPE).
This result highlights the importance of conditioning the deformation field on a latent representation of organ state.
Removing canonicalisation produces the largest drop in tracking accuracy and also substantially degrades reconstruction (mean error rises from 0.86/1.47 to 0.96/4.23 (CD/EPE)), confirming that aligning organs to a shared canonical space stabilises deformation learning across timesteps.
Replacing our latent conditioning with Feature-wise Linear Modulation (FiLM) conditioning \cite{perez_film_2018}, which conditions the network by modulating intermediate feature activations through learned feature-wise affine transformations, similarly reduces performance (0.96/2.47).
This suggests that directly injecting the latent code provides a stronger inductive bias for modelling plant motion than conditioning through feature modulation alone.
Finally, removing the multi-step Chamfer loss leads to moderate but consistent degradation (0.91/1.89), indicating that it improves temporal stability and geometric consistency.
Overall, the ablation results confirm that latent conditioning and canonicalisation are the most important components of the proposed formulation.

\begin{table}[tb]
\caption{\textbf{Ablations results and evaluation on full plants.} Removing latent conditioning and canonicalisation leads to the most noticeable drop in reconstruction and tracking accuracy, highlighting the importance of these design choices. *Note that \textit{w/model segmentation} is not directly comparable to results with GT segmentation given different averaging or number of leaf-tip evaluations (only provided as reference values). All metrics are averaged over five independent runs.}
\centering
\label{tab:ablations}
\resizebox{\textwidth}{!}{
\begin{tabular}{l cc cc cc cc cc cc cc cc cc}
\toprule
 & \multicolumn{2}{c}{MaC2} & \multicolumn{2}{c}{MaC3} & \multicolumn{2}{c}{SoC2} & \multicolumn{2}{c}{SoH2} & \multicolumn{2}{c}{TbC1} & \multicolumn{2}{c}{TbS3} & \multicolumn{2}{c}{Tm1H3} & \multicolumn{2}{c}{Tm1S1} & \multicolumn{2}{c}{\textit{Mean}} \\
\cmidrule(lr){2-3} \cmidrule(lr){4-5} \cmidrule(lr){6-7} \cmidrule(lr){8-9} \cmidrule(lr){10-11} \cmidrule(lr){12-13} \cmidrule(lr){14-15} \cmidrule(lr){16-17} \cmidrule(lr){18-19}
Method & CD $\downarrow$ & EPE $\downarrow$ & CD $\downarrow$ & EPE $\downarrow$ & CD $\downarrow$ & EPE $\downarrow$ & CD $\downarrow$ & EPE $\downarrow$ & CD $\downarrow$ & EPE $\downarrow$ & CD $\downarrow$ & EPE $\downarrow$ & CD $\downarrow$ & EPE $\downarrow$ & CD $\downarrow$ & EPE $\downarrow$ & CD $\downarrow$ & EPE $\downarrow$ \\
\midrule
\textbf{Ours (full)} & \cellcolor{best} 2.51 & \cellcolor{best} 2.38 & \cellcolor{best} 1.43 & \cellcolor{second} 2.01 & \cellcolor{second} 0.69 & \cellcolor{best} 1.34 & \cellcolor{best} 0.74 & \cellcolor{best} 0.97 & \cellcolor{second} 0.59 & \cellcolor{best} 1.66 & \cellcolor{best} 0.45 & \cellcolor{second} 1.30 & \cellcolor{best} 0.12 & \cellcolor{best} 0.80 & \cellcolor{best} 0.32 & \cellcolor{best} 1.34 & \cellcolor{best} 0.86 & \cellcolor{best} 1.47 \\
\midrule
\quad \textit{w/o canonicalisation} & 2.76 & 3.14 & 1.58 & 2.28 & 0.93 & 3.41 & \cellcolor{third} 0.79 & 13.24 & \cellcolor{third} 0.60 & 5.79 & \cellcolor{third} 0.47 & \cellcolor{third} 1.40 & 0.21 & 3.18 & \cellcolor{third} 0.34 & \cellcolor{third} 1.42 & 0.96 & 4.23 \\
\quad \textit{w/o latent code} & 2.77 & 8.04 & 1.76 & 3.70 & 0.83 & 1.43 & 1.00 & 3.32 & 0.83 & 3.72 & 0.61 & 3.31 & 0.15 & 1.28 & 0.47 & 2.72 & 1.05 & 3.44 \\
\quad \textit{w/o multi-step CD loss} & 2.87 & \cellcolor{third} 2.79 & \cellcolor{second} 1.45 & 2.21 & \cellcolor{best} 0.68 & \cellcolor{second} 1.37 & \cellcolor{second} 0.75 & \cellcolor{third} 1.04 & \cellcolor{best} 0.58 & \cellcolor{third} 3.29 & \cellcolor{second} 0.46 & 1.65 & \cellcolor{third} 0.14 & 1.34 & 0.38 & 1.48 & \cellcolor{third} 0.91 & \cellcolor{third} 1.89 \\
\quad \textit{w/o $\mathcal{L}_{code}$} & \cellcolor{second} 2.52 & \cellcolor{second} 2.46 & \cellcolor{third} 1.54 & \cellcolor{third} 2.14 & \cellcolor{third} 0.71 & \cellcolor{third} 1.38 & \cellcolor{best} 0.74 & \cellcolor{second} 0.99 & 0.61 & 3.42 & \cellcolor{best} 0.45 & \cellcolor{best} 1.06 & \cellcolor{best} 0.12 & \cellcolor{second} 0.85 & \cellcolor{best} 0.32 & 1.62 & \cellcolor{second} 0.88 & \cellcolor{second} 1.74 \\
\quad \textit{w/ FiLM conditioning \cite{perez_film_2018}} & \cellcolor{third} 2.62 & 5.74 & 1.67 & \cellcolor{best} 1.93 & 0.83 & 1.39 & 0.92 & 3.32 & 0.67 & \cellcolor{second} 3.04 & \cellcolor{third} 0.47 & 2.04 & \cellcolor{second} 0.13 & \cellcolor{third} 0.92 & \cellcolor{second} 0.33 & \cellcolor{second} 1.35 & 0.96 & 2.47 \\
\midrule
\quad \textit{w/ model segmentation}* & 1.77 & 2.17 & 1.21 & 1.94 & 0.68 & 1.32 & 0.48 & 0.80 & 2.01 & 1.59 & 0.36 & 1.55 & 0.12 & 0.61 & 0.27 & 1.36 & 0.86 & 1.42 \\

\bottomrule
\end{tabular}
}
\end{table}

\textbf{Robustness.}
Our method assumes organ-segmented inputs, a reasonable requirement given the rapidly maturing literature on automated 3D plant organ segmentation, which is already of sufficient quality for organ tracking~\cite{fusaro_horticultural_2026, li_trackplant3d_2024,  lobefaro_spatio-temporal_2025, schunck_pheno4d_2021}.
To verify that GrowFields does not rely on ground-truth labels, we re-run it on \emph{automatically} segmented point clouds (\textit{w/ model segmentation} in \autoref{tab:ablations}), obtained with a combination of a retrained PSegNet \cite{li_psegnet_2022} model on all available sequences, and the TrackPlant3D organ tracking method \cite{li_trackplant3d_2024}.
As illustrated qualitatively in \autoref{fig:autoseg_qual}, the method remains effective under realistic preprocessing, yielding a mean CD~0.86 / EPE~1.42 versus 0.86 / 1.47 with ground-truth labels.
We further probe robustness to imperfect geometry by adding Gaussian noise and by removing local point neighbourhoods (\emph{holes}) around test regions.
GrowFields degrades gracefully in both cases; notably, the holes experiment confirms it learns genuine growth dynamics rather than exploiting nearest-neighbour train--test proximity, with full details in the supplementary material.

\textbf{Statistical significance.}
Given the limited number of full-plant sequences, we assess significance at the organ level, considering the results of \autoref{tab:results_per_organ} ($45$ organ sequences in total).
A Wilcoxon signed-rank test on per-organ metrics (Shapiro–Wilk $p\!<\!0.001$, indicating non-normal distributions) shows that GrowFields significantly outperforms COAP on both CD and EPE (both $p\!<\!0.001$), and DPF on CD ($p\!<\!0.001$) and EPE ($p\!=\!0.004$).

\section{Conclusion}
\label{sec:conclusion}

We introduced GrowFields, a compositional neural field framework for modelling 4D plant growth from sparse longitudinal point clouds. 
By canonicalising plant organs and learning a shared latent-conditioned deformation field, the model captures asynchronous organ development and topology changes within a unified representation. 
Experiments across multiple plant species show that this formulation improves both geometric reconstruction and organ-level tracking compared to existing dynamic reconstruction methods. 
Our evaluation protocol based on annotated leaf-tip trajectories further provides a practical benchmark for studying plant growth dynamics.
While we focus on plants, the formulation is general and could apply to other dynamic point clouds composed of independently evolving parts, such as 4D cell or vessel growth in the medical domain~\cite{wiesner_generative_2024}.
We hope this work contributes a useful representation for learning structure-aware models of plant development and enables future research on data-driven plant phenotyping and analysis.

\subsection{Limitations and outlook}
\label{sec:limitations}
Our approach has three main limitations.
First, because we warp the initial organ point cloud forward without adding points, regions undergoing rapid growth become spatially sparse relative to the ground-truth scans (\eg the leaf at day~7 in \autoref{fig:fullplant_methods}); a promising direction is adaptive densification analogous to 3DGS~\cite{kerbl3Dgaussians} and 4DGS~\cite{Wu_2024_CVPR}.
Second, very short organ sequences ($\leq$5 training timesteps) leave the velocity field underconstrained, as reflected in the extrapolation experiment and the SoH2 sequence.
Third, as noted in \autoref{sec:method}, GrowFields does not predict organ emergence; newly observed organs are only incorporated upon first observation.
Incorporating priors from L-systems or generative models to anticipate morphogenesis is an interesting avenue for future work \cite{Ghrer2026_cvpr}.
Beyond these limitations, given the increasing availability of 4D plant point-cloud datasets, a further promising direction is to harmonise and combine multiple datasets to learn generative geometric models of 4D plant growth that generalise across species and environmental conditions \cite{Ahmed2026_Pepper4D, drees_time_2022}.

\section*{Acknowledgements} 
This work was supported by funding from the \textit{Smart Sustainable Farming Research Program} from the World Food System Center from ETH Zürich.

%
%
\bibliographystyle{splncs04}
\bibliography{main}

\newpage
\clearpage
\setcounter{page}{1}
\section*{Supplementary Material}

\section*{Dataset details}

The sequence name abbreviations used in the paper are defined below. Each sequence corresponds to a time series of 3D plant scans acquired over multiple days under different stress conditions. The processed dataset, as well as the code and trained models, will be made publicly available upon publication.
\begin{small}
\begin{verbatim}
{
  "MaC2": "maize_control_plant2",
  "MaC3": "maize_control_plant3",
  "SoC2": "sorghum_control_plant2",
  "SoH2": "sorghum_highlight_plant2",
  "TbC1": "tobacco_control_plant1",
  "TbS3": "tobacco_shade_plant3",
  "Tm1H3": "tomato1_heat_plant3",
  "Tm1S1": "tomato1_shade_plant1"
}
\end{verbatim}
\end{small}
The number of timepoints per sequence is: MaC2 (9), MaC3 (8), SoC2 (8), SoH2 (5), TbC1 (7), TbS3 (8), Tm1H3 (16), and Tm1S1 (17).

\subsubsection*{Dataset construction.}
The TrackPlant3D dataset~\cite{li_trackplant3d_2024} comprises 43 daily LiDAR acquisitions from four plant species (maize, sorghum, tobacco, and tomato), collected from two publicly available sources~\cite{schunck_pheno4d_2021, Conn_2017_lidar}.
Each scan contains approximately 10,000 evenly spaced points per plant, with semantic instance labels identifying each organ (0 for the stem and positive integers for individual leaves).
To make the dataset suitable for evaluating 4D plant reconstruction, we annotated leaf tips on all scans where they could be clearly identified, yielding 234 leaf-tip points in total.
We selected a subset of three plants per species, chosen a posteriori based on the highest availability of leaf-tip annotations across the sequence, and divided them into a validation set of four sequences (one per species, for hyperparameter tuning) and a test set of eight sequences used to assess generalisation.
All plant sequences contain between 5 and 17 timepoints, and individual organ sequences range from 1 to 17 (for \autoref{tab:results_per_organ} organs sequences with less than 3 timesteps were not considered).

\subsubsection*{Implementation details.}
Our 4D SIREN model is a multi-layer perceptron with five layers of 256 units, trained for 50,000 steps for each plant sequence using Adam~\cite{kingma2015adam} with an initial learning rate of $5e^{-6}$, halved at plateau with 10 steps patience.
We use latent codes of size 32, initialised with a variance of 0.1 and optimised with an initial learning rate of $5e^{-3}$.
We set $\lambda_{code}=1e^{-4}$.
At each training step we sample consecutive frames 50\% of the time and randomly sample 8192 points from the source frame, with an equal number of points per organ in the mini-batch.
We use a mini-batch size of 10, found through grid search on validation sequences.
Code and trained models will be made publicly available upon publication.

\subsubsection*{Baselines.}
The full-plant baselines are CanFields~\cite{Wang_2025_ICCV}, NVFi~\cite{li_2023_neurips}, NDF~\cite{Sun_2022_CVPR}, and the 4D SDF model DSR~\cite{Sun_2023_neurips}, none of which take segmentation as input.
As a point-based baseline we use chained Dynamic Point Fields (DPF)~\cite{Prokudin_2023_ICCV}, modelling each frame transition with a separate MLP, and additionally evaluate it on a per-organ basis.
As a compositional, part-aware baseline we extend the COAP~\cite{mihajlovic_coap_2022} encoder-decoder architecture into a dynamic deformation field trained in the same way as GrowFields, using the same ground-truth organ segmentation as our method.



\section*{Qualitative interpolation results}
\begin{figure}
\centering
\setlength{\tabcolsep}{2pt}

\begin{tikzpicture}
\node (table) at (0,0) {
\begin{tabular}{m{0.9cm} m{0.95\linewidth}}
    \rotatebox{90}{\textbf{COAP}~\cite{mihajlovic_coap_2022}} &
    \begin{tikzpicture}
    \node[anchor=south west,inner sep=0] (img) at (0,0) {
    \includegraphics[width=0.99\linewidth, trim={7cm 5cm 7cm 4cm}, clip]
    {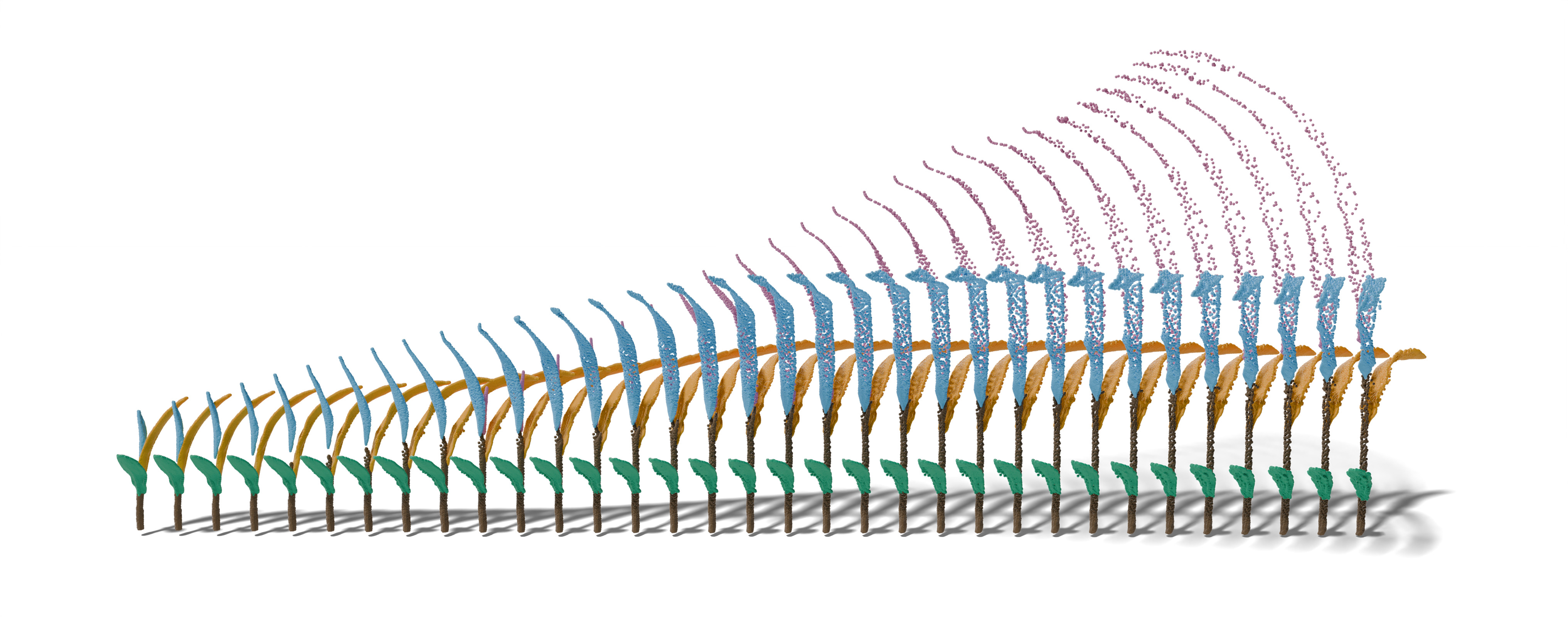}
    };
    \begin{scope}[x={(img.south east)},y={(img.north west)}]
    \draw[red!80!black, line width=0.8pt] (0.79,0.98) ellipse (0.04 and 0.05);
    \draw[red!80!black, line width=0.8pt] (0.913,0.33) ellipse (0.02 and 0.1);
    \end{scope}
    \end{tikzpicture} \\

    \rotatebox{90}{\textbf{DPF}~\cite{Prokudin_2023_ICCV} \textbf{(per-organ)}} &
    \begin{tikzpicture}
    \node[anchor=south west,inner sep=0] (img) at (0,0) {
    \includegraphics[width=0.99\linewidth, trim={7cm 5cm 7cm 4cm}, clip]
    {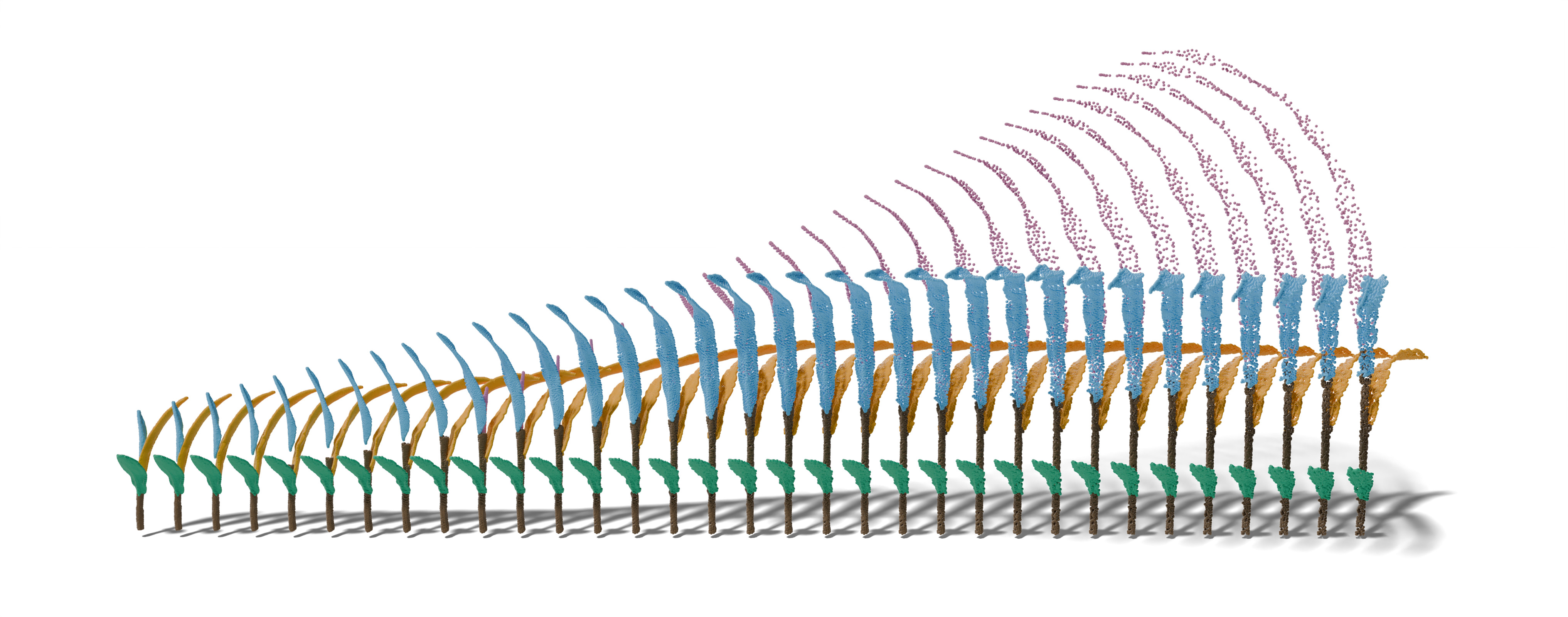}
    };
    \begin{scope}[x={(img.south east)},y={(img.north west)}]
    \draw[red!80!black, line width=0.8pt] (0.79,0.98) ellipse (0.04 and 0.05);
    \draw[red!80!black, line width=0.8pt] (0.913,0.33) ellipse (0.02 and 0.1);
    \end{scope}
    \end{tikzpicture}\\

    \rotatebox{90}{\textbf{Ours}} &
    \begin{tikzpicture}
    \node[anchor=south west,inner sep=0] (img) at (0,0) {
    \includegraphics[width=0.99\linewidth, trim={7cm 5cm 7cm 4cm}, clip]
    {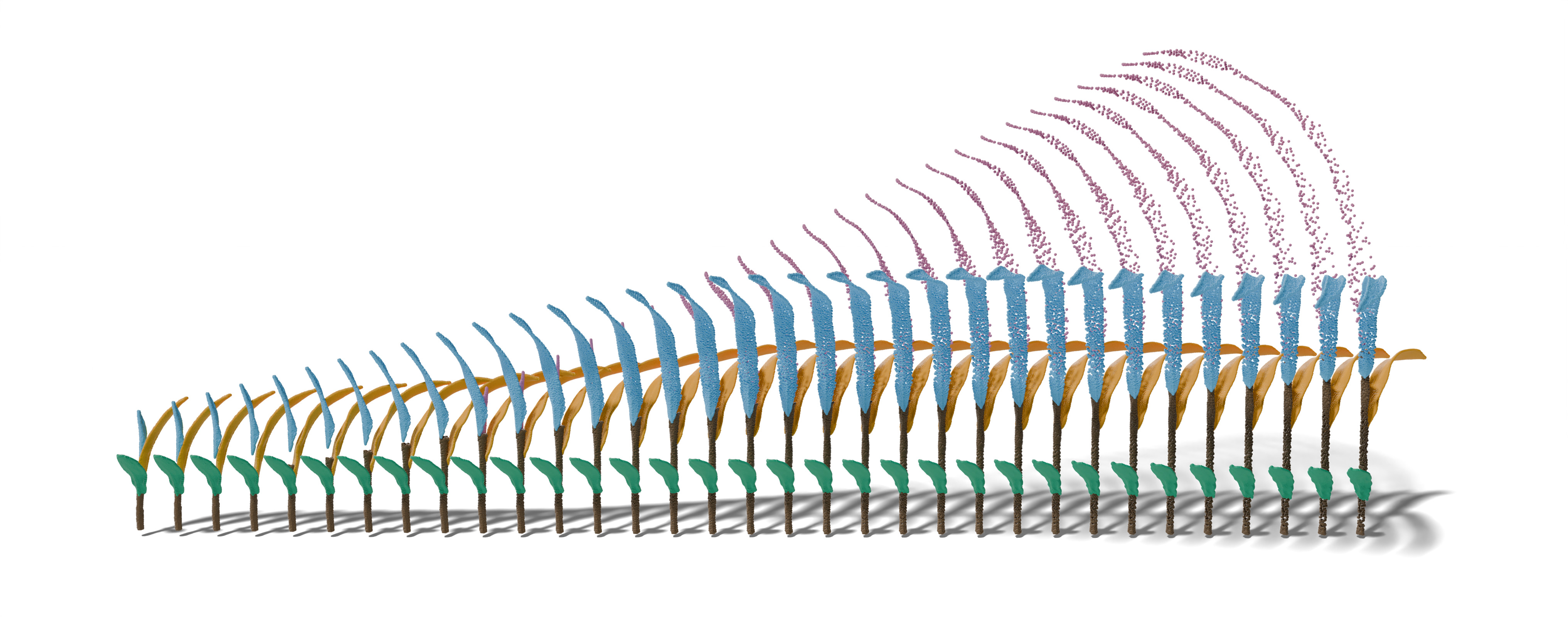}
    };
    \begin{scope}[x={(img.south east)},y={(img.north west)}]
    \draw[green!60!black, line width=0.8pt] (0.79,0.98) ellipse (0.04 and 0.05);
    \draw[green!60!black, line width=0.8pt] (0.913,0.33) ellipse (0.02 and 0.1);
    \end{scope}
    \end{tikzpicture}\\

    \rotatebox{90}{\textbf{GT}} &
    \begin{tikzpicture}
    \node[anchor=south west,inner sep=0] (img) at (0,0) {
    \includegraphics[width=0.99\linewidth, trim={7cm 5cm 7cm 4cm}, clip]
    {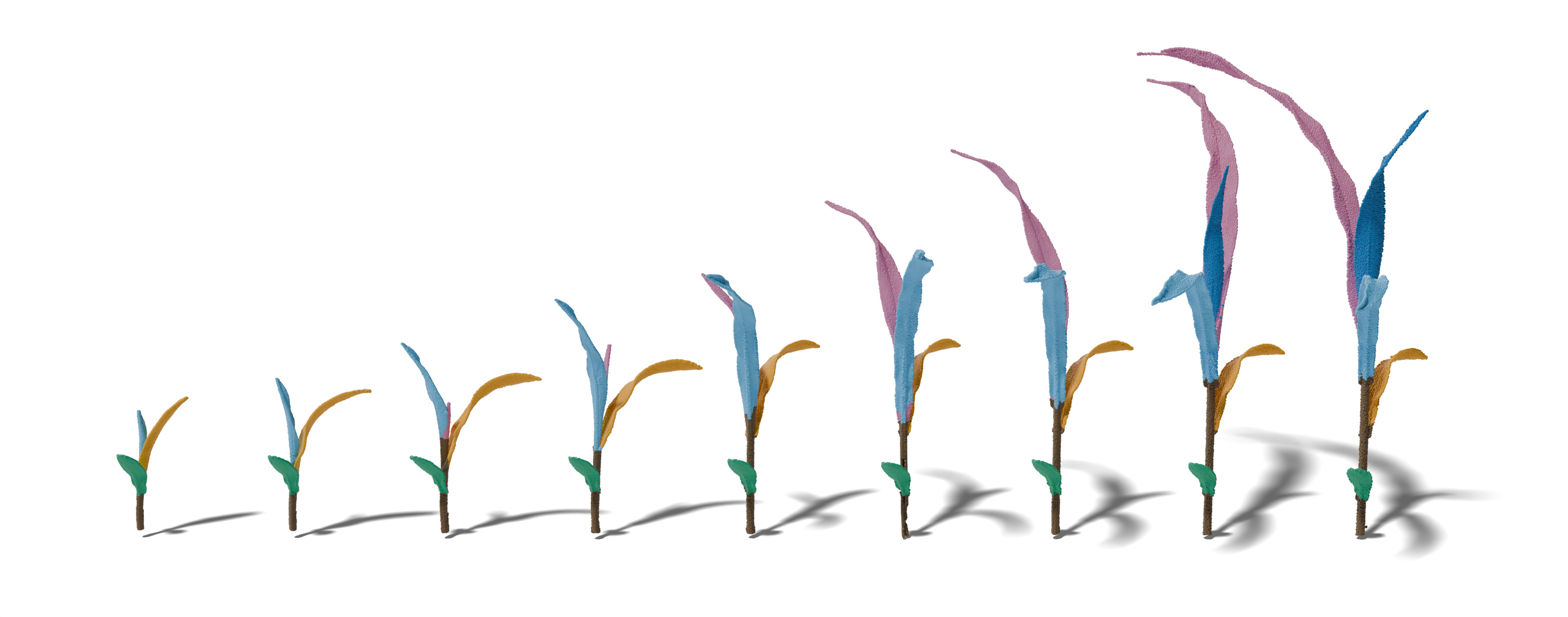}
    };
    \def\leftmargin{0.045}
    \def\rightmargin{0.9075}
    \foreach \i in {0,...,8} {
    \pgfmathsetmacro\t{\leftmargin + (\rightmargin-\leftmargin)*\i/8}
    \path ($(img.south west)!\t!(img.south east)$) coordinate (p);
    \ifodd\i
        \def\col{gray}\def\yshift{-0.25}
    \else
        \def\col{black}\def\yshift{0.0}
    \fi
    \node[\col,font=\scriptsize\bfseries,anchor=north] at ($(p)+(0,\yshift)$) {Day \i};
    }
    \end{tikzpicture}
    
    \end{tabular}
    };
    
    \begin{scope}[overlay,
        x={($(table.south east)-(table.south west)$)},
        y={($(table.north west)-(table.south west)$)},
        shift={(table.south west)}]
    \def\leftmargin{0.135}
    \def\rightmargin{0.895}
    \foreach \i in {0,...,8}{
        \pgfmathsetmacro\t{\leftmargin + (\rightmargin-\leftmargin)*\i/8}
        \ifodd\i
            \def\linecol{gray!20}
        \else
            \def\linecol{black!30}
        \fi
        \draw[\linecol, dashed, line width=0.2pt]
            (\t,0.1) -- (\t,0.8);
    }
    \end{scope}
    
    \end{tikzpicture}
        
    \vspace{-25pt}
    \caption{Dense interpolation on \texttt{maize\_control\_plant2} sequence. \textbf{Black} labels indicate training frames, while \textcolor{gray}{\textbf{grey}} labels denote evaluation frames. Our method produces smoother organ joints, sharper leaf boundaries, and more consistent deformation fields.}
    \label{fig:interp_comparison_maize_control_plant2}

\end{figure}
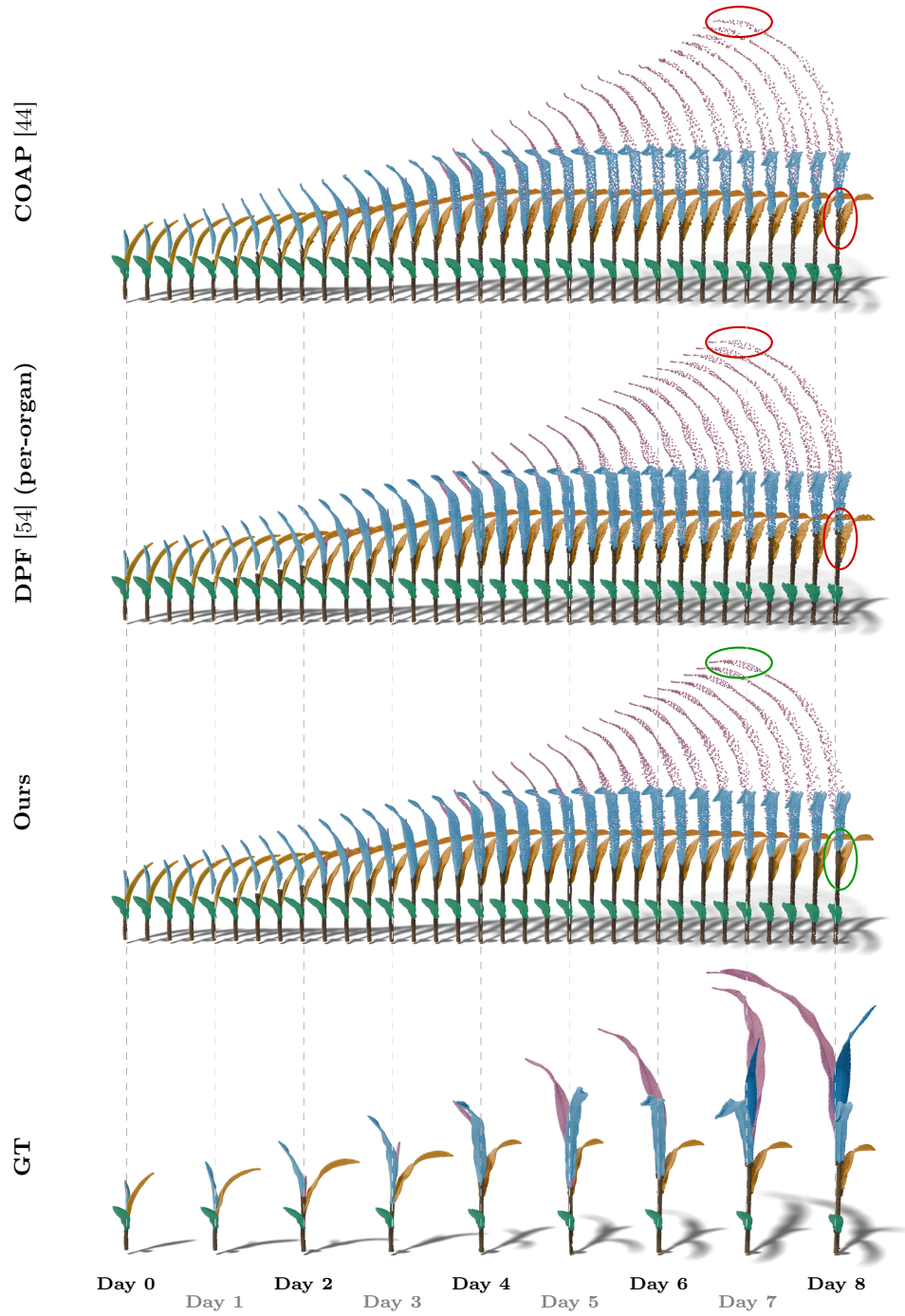

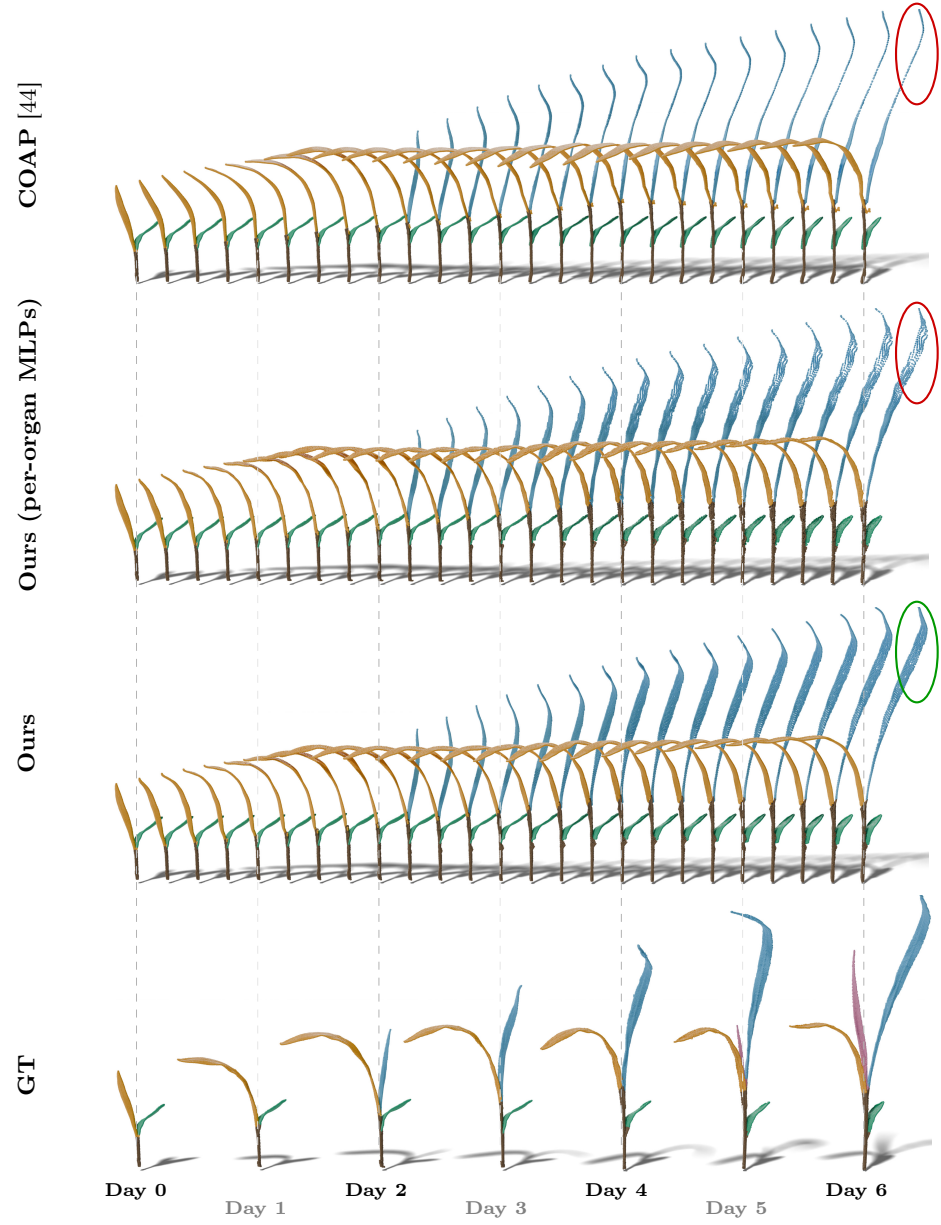
\begin{figure}
\centering
\setlength{\tabcolsep}{2pt}

\begin{tikzpicture}
\node (table) at (0,0) {
\begin{tabular}{m{0.9cm} m{0.95\linewidth}}
    \rotatebox{90}{\textbf{COAP}~\cite{mihajlovic_coap_2022}} &
    \begin{tikzpicture}
    \node[anchor=south west,inner sep=0] (img) at (0,0) {
    \includegraphics[width=0.95\linewidth, trim={7cm 7cm 7cm 7cm}, clip]
    {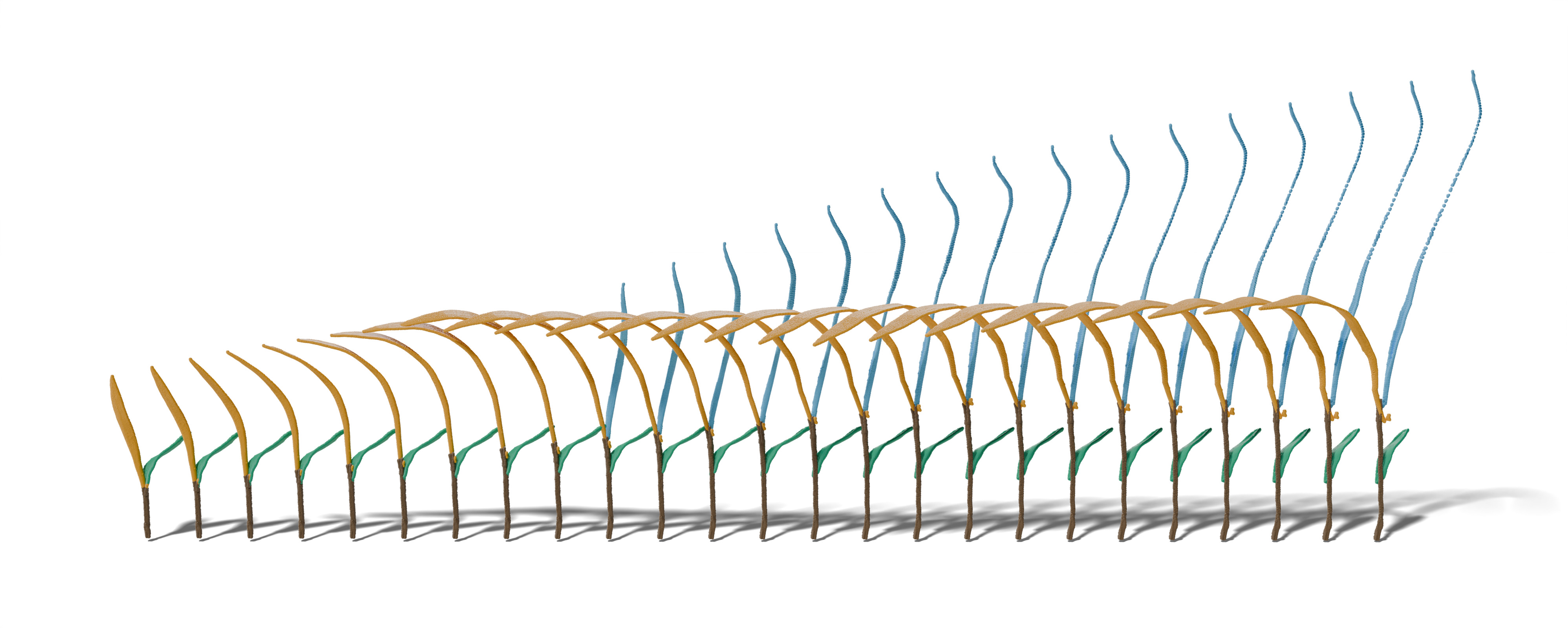}
    };
    \begin{scope}[x={(img.south east)},y={(img.north west)}]
    \draw[red!80!black, line width=0.8pt] (0.985,0.84) ellipse (0.025 and 0.18);
    \end{scope}
    \end{tikzpicture}\\
    
    \rotatebox{90}{\textbf{Ours (per-organ MLPs)}} &
    \begin{tikzpicture}
    \node[anchor=south west,inner sep=0] (img) at (0,0) {
    \includegraphics[width=0.95\linewidth, trim={7cm 7cm 7cm 7cm}, clip]
    {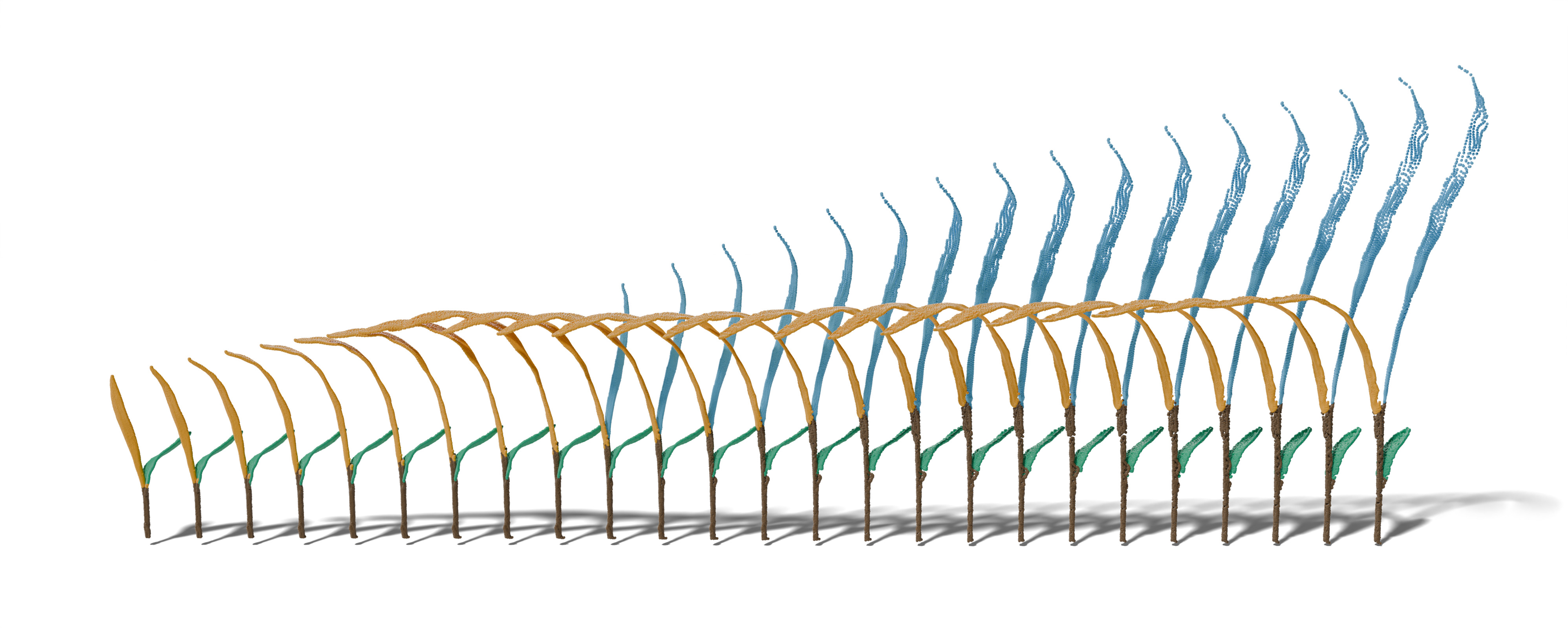}
    };
    \begin{scope}[x={(img.south east)},y={(img.north west)}]
    \draw[red!80!black, line width=0.8pt] (0.985,0.84) ellipse (0.025 and 0.18);
    \end{scope}
    \end{tikzpicture}\\

    \rotatebox{90}{\textbf{Ours}} &
    \begin{tikzpicture}
    \node[anchor=south west,inner sep=0] (img) at (0,0) {
    \includegraphics[width=0.95\linewidth, trim={7cm 7cm 7cm 7cm}, clip]
    {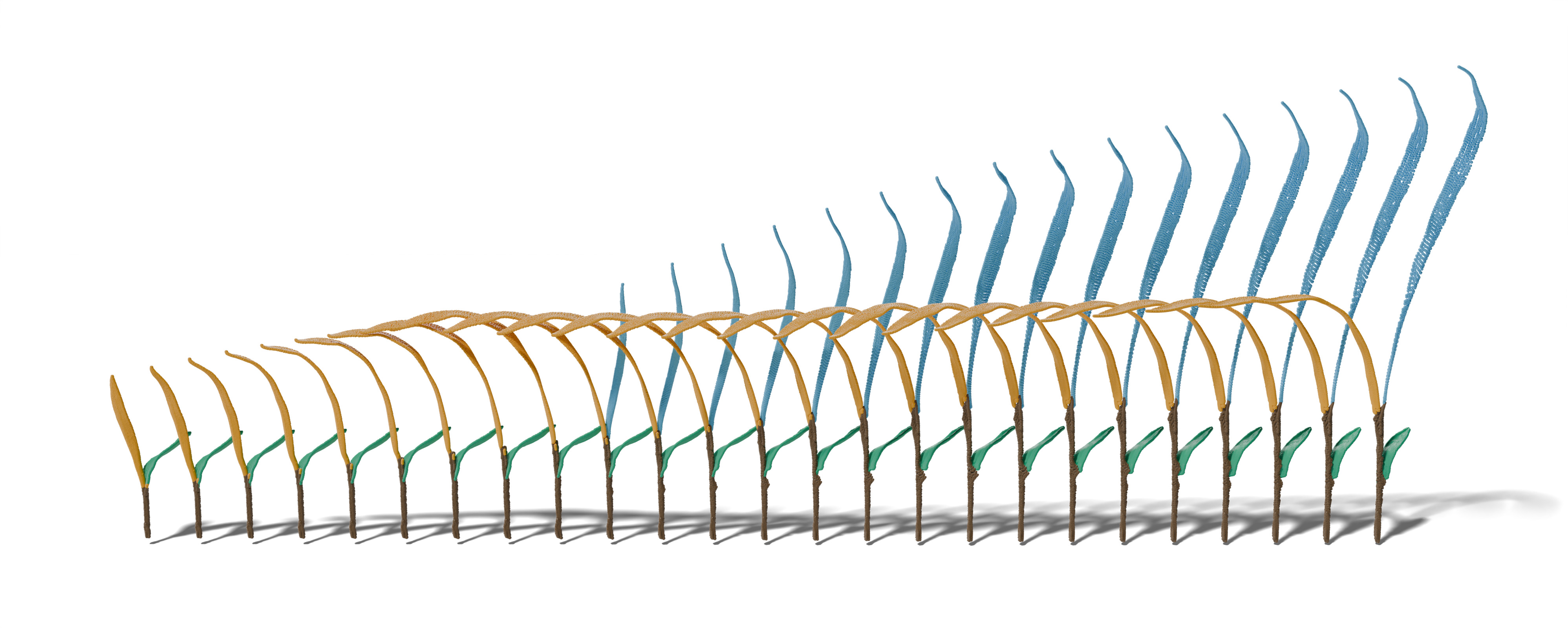}
    };
    \begin{scope}[x={(img.south east)},y={(img.north west)}]
    \draw[green!60!black, line width=0.8pt] (0.985,0.84) ellipse (0.025 and 0.18);
    \end{scope}
    \end{tikzpicture}\\

    \rotatebox{90}{\textbf{GT}} &
    \begin{tikzpicture}
    \node[anchor=south west,inner sep=0] (img) at (0,0) {
    \includegraphics[width=0.95\linewidth, trim={7cm 7cm 7cm 7cm}, clip]
    {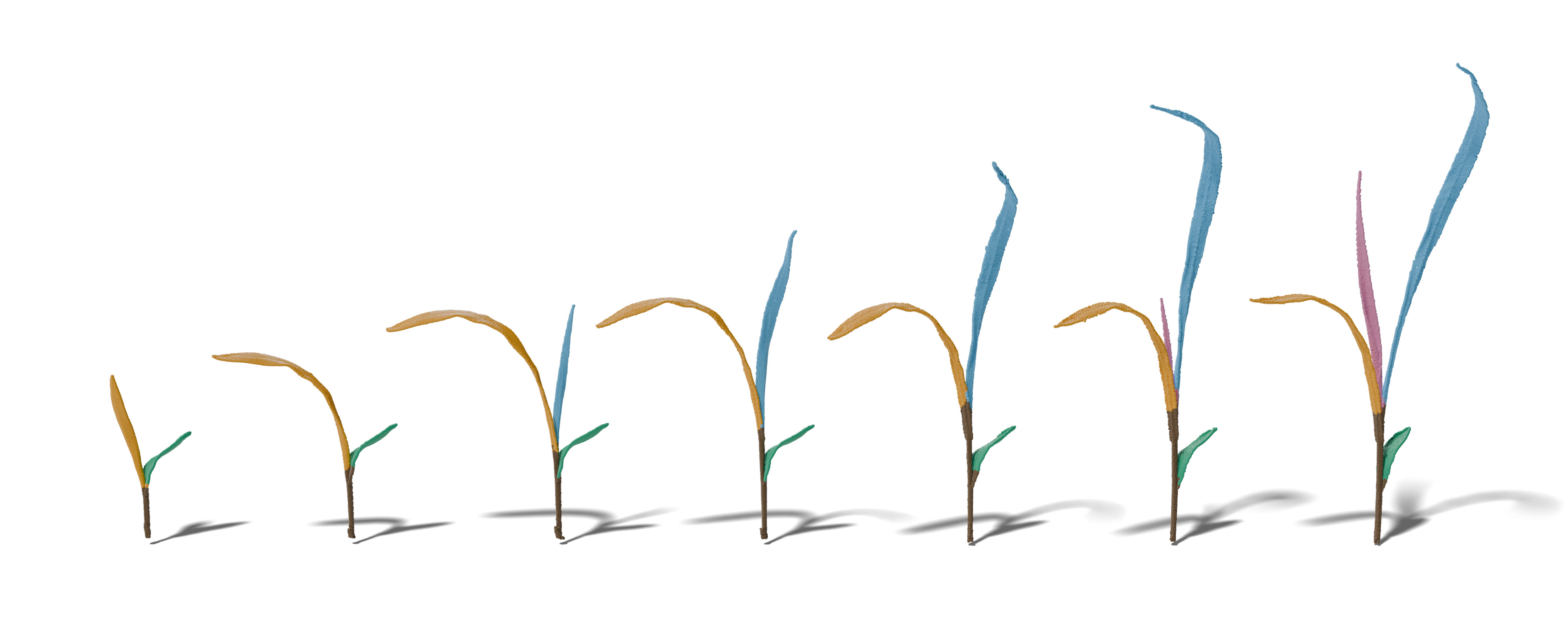}
    };
    \def\leftmargin{0.045}
    \def\rightmargin{0.91}
    \foreach \i in {0,...,6} {
    \pgfmathsetmacro\t{\leftmargin + (\rightmargin-\leftmargin)*\i/6}
    \pgfmathtruncatemacro{\day}{2 + 2*\i}
    \path ($(img.south west)!\t!(img.south east)$) coordinate (p);
    \ifodd\i
        \def\col{gray}\def\yshift{-0.25}
    \else
        \def\col{black}\def\yshift{0.0}
    \fi
    \node[\col,font=\scriptsize\bfseries,anchor=north] at ($(p)+(0,\yshift)$) {Day \i};
    }
    \end{tikzpicture}
    
    \end{tabular}
    };
    
    \begin{scope}[overlay,
        x={($(table.south east)-(table.south west)$)},
        y={($(table.north west)-(table.south west)$)},
        shift={(table.south west)}]
    \def\leftmargin{0.135}
    \def\rightmargin{0.875}
    \foreach \i in {0,...,6}{
        \pgfmathsetmacro\t{\leftmargin + (\rightmargin-\leftmargin)*\i/6}
        \ifodd\i
            \def\linecol{gray!20}
        \else
            \def\linecol{black!30}
        \fi
        \draw[\linecol, dashed, line width=0.2pt]
            (\t,0.1) -- (\t,0.8);
    }
    \end{scope}
    
    \end{tikzpicture}
    
    \vspace{-25pt}
    \caption{Dense interpolation on \texttt{sorghum\_control\_plant2} sequence. \textbf{Black} labels indicate training frames, while \textcolor{gray}{\textbf{grey}} labels denote evaluation frames. Our method preserves leaf shape and produces more consistent deformation fields.}
    \label{fig:interp_comparison_sorghum_control_plant2}

\end{figure}

\begin{figure}
\centering
\setlength{\tabcolsep}{2pt}

\begin{tikzpicture}
\node (table) at (0,0) {
\begin{tabular}{m{0.9cm} m{0.95\linewidth}}
    \rotatebox{90}{\textbf{COAP}~\cite{mihajlovic_coap_2022}} &
    \begin{tikzpicture}
    \node[anchor=south west,inner sep=0] (img) at (0,0) {
    \includegraphics[width=0.99\linewidth, trim={7cm 5cm 7cm 4cm}, clip]
    {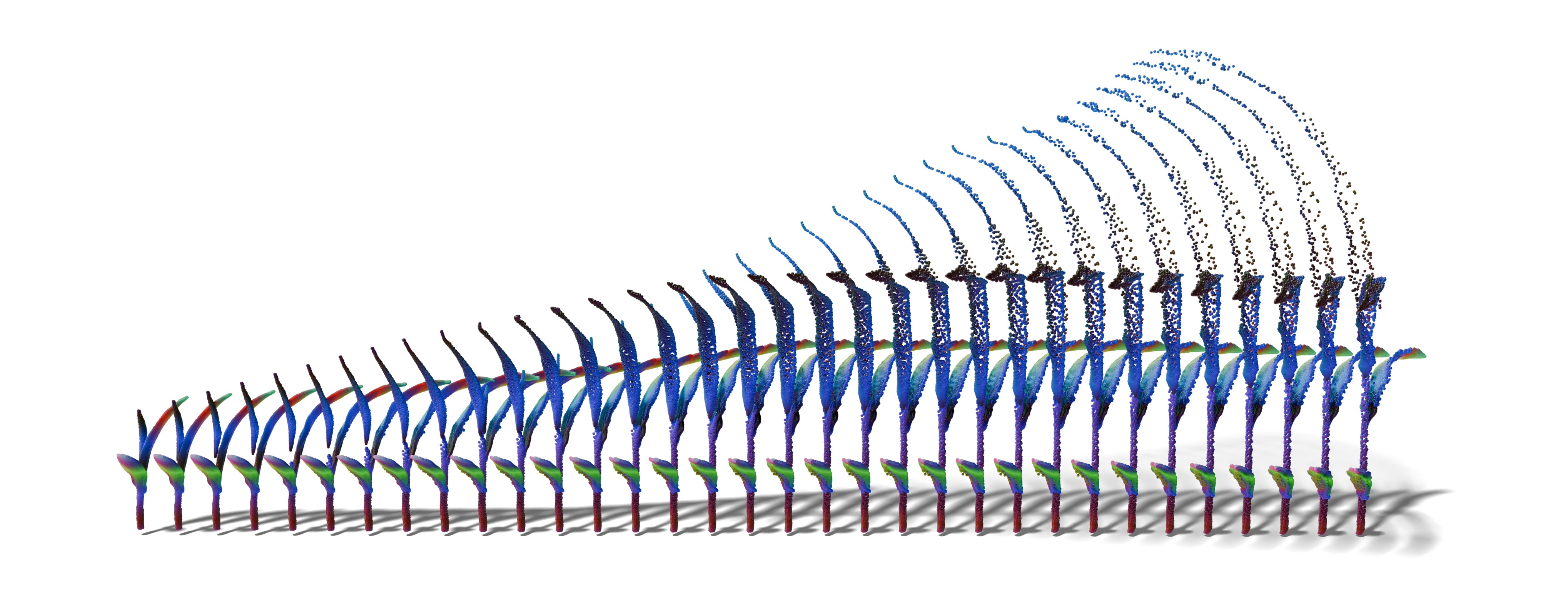}
    };
    \begin{scope}[x={(img.south east)},y={(img.north west)}]
    \draw[red!80!black, line width=0.8pt] (0.933,0.4) ellipse (0.03 and 0.05);
    \end{scope}
    \end{tikzpicture} \\

    \rotatebox{90}{\textbf{DPF}~\cite{Prokudin_2023_ICCV} \textbf{(per-organ)}} &
    \begin{tikzpicture}
    \node[anchor=south west,inner sep=0] (img) at (0,0) {
    \includegraphics[width=0.99\linewidth, trim={7cm 5cm 7cm 4cm}, clip]
    {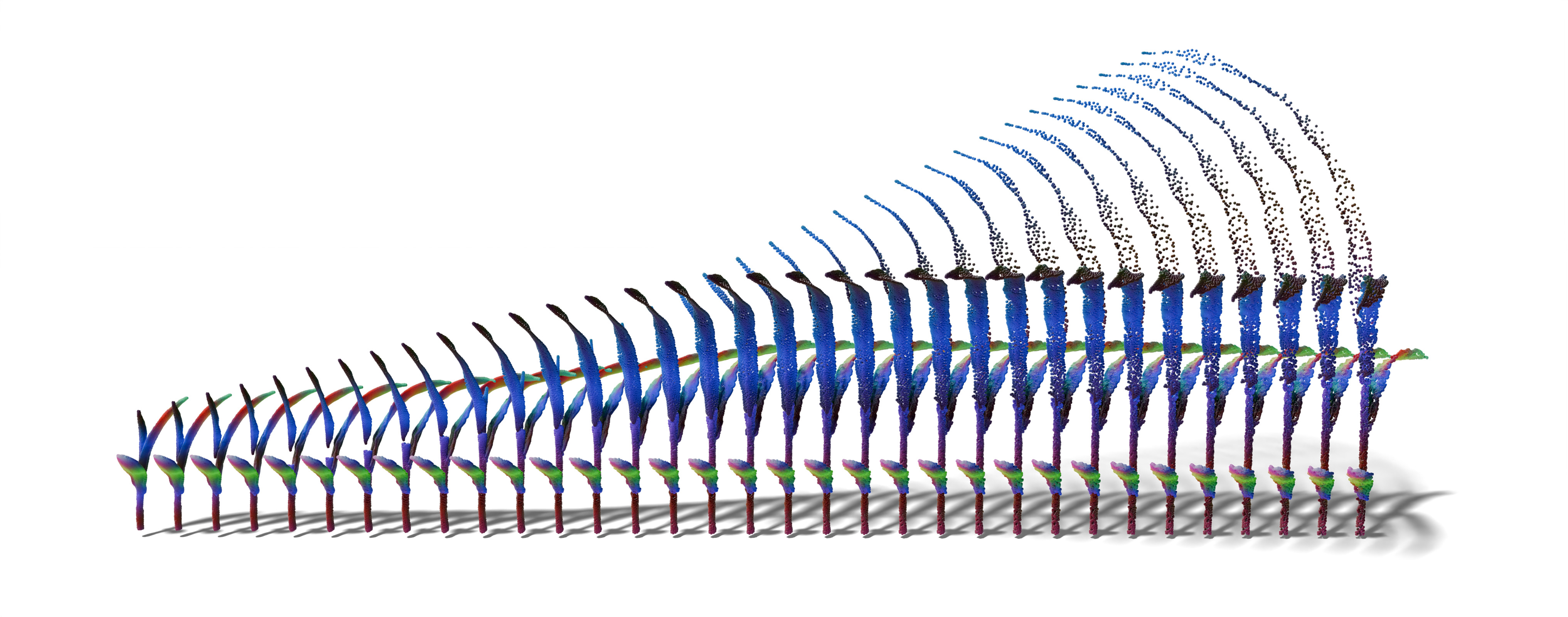}
    };
    \begin{scope}[x={(img.south east)},y={(img.north west)}]
    \draw[red!80!black, line width=0.8pt] (0.933,0.4) ellipse (0.03 and 0.05);
    \end{scope}
    \end{tikzpicture}\\

    \rotatebox{90}{\textbf{Ours (per-organ MLPs)}} &
    \begin{tikzpicture}
    \node[anchor=south west,inner sep=0] (img) at (0,0) {
    \includegraphics[width=0.99\linewidth, trim={7cm 5cm 7cm 4cm}, clip]
    {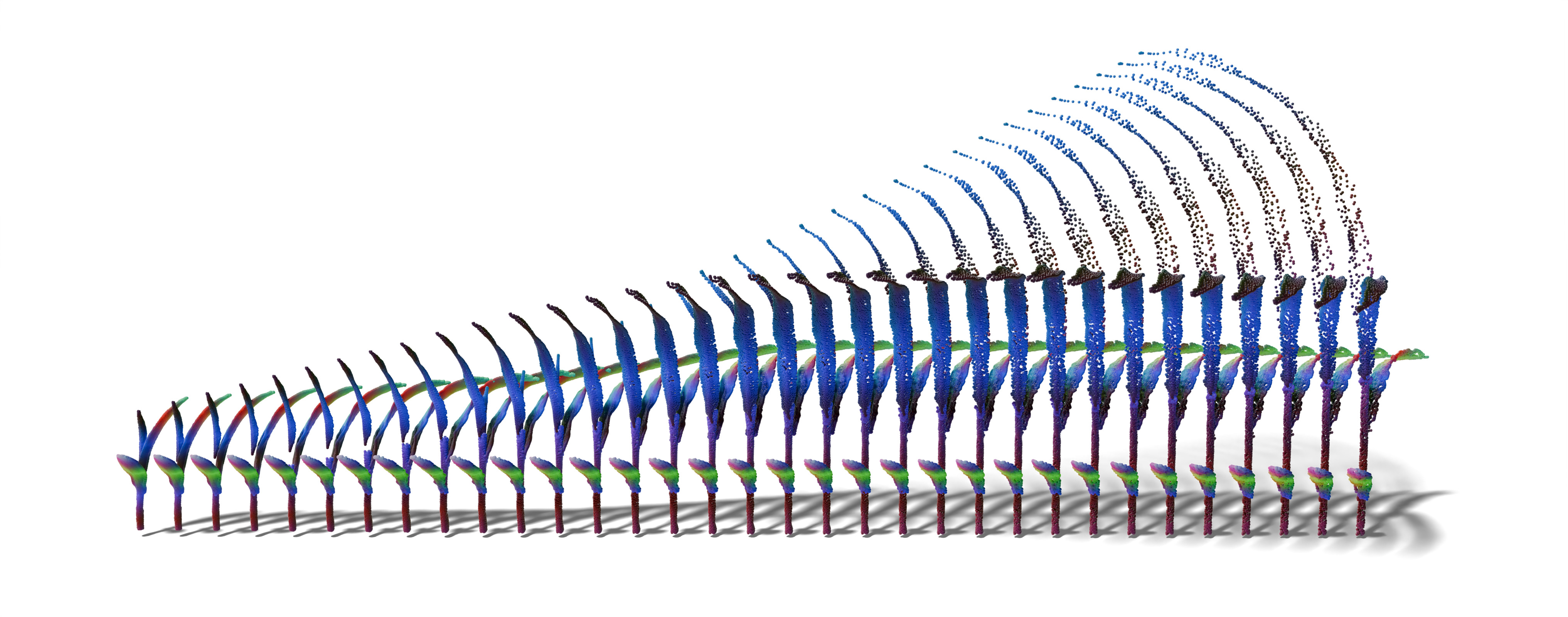}
    };
    \begin{scope}[x={(img.south east)},y={(img.north west)}]
    \draw[red!80!black, line width=0.8pt] (0.933,0.4) ellipse (0.03 and 0.05);
    \end{scope}
    \end{tikzpicture}\\

    \rotatebox{90}{\textbf{Ours}} &
    \begin{tikzpicture}
    \node[anchor=south west,inner sep=0] (img) at (0,0) {
    \includegraphics[width=0.99\linewidth, trim={7cm 5cm 7cm 4cm}, clip]
    {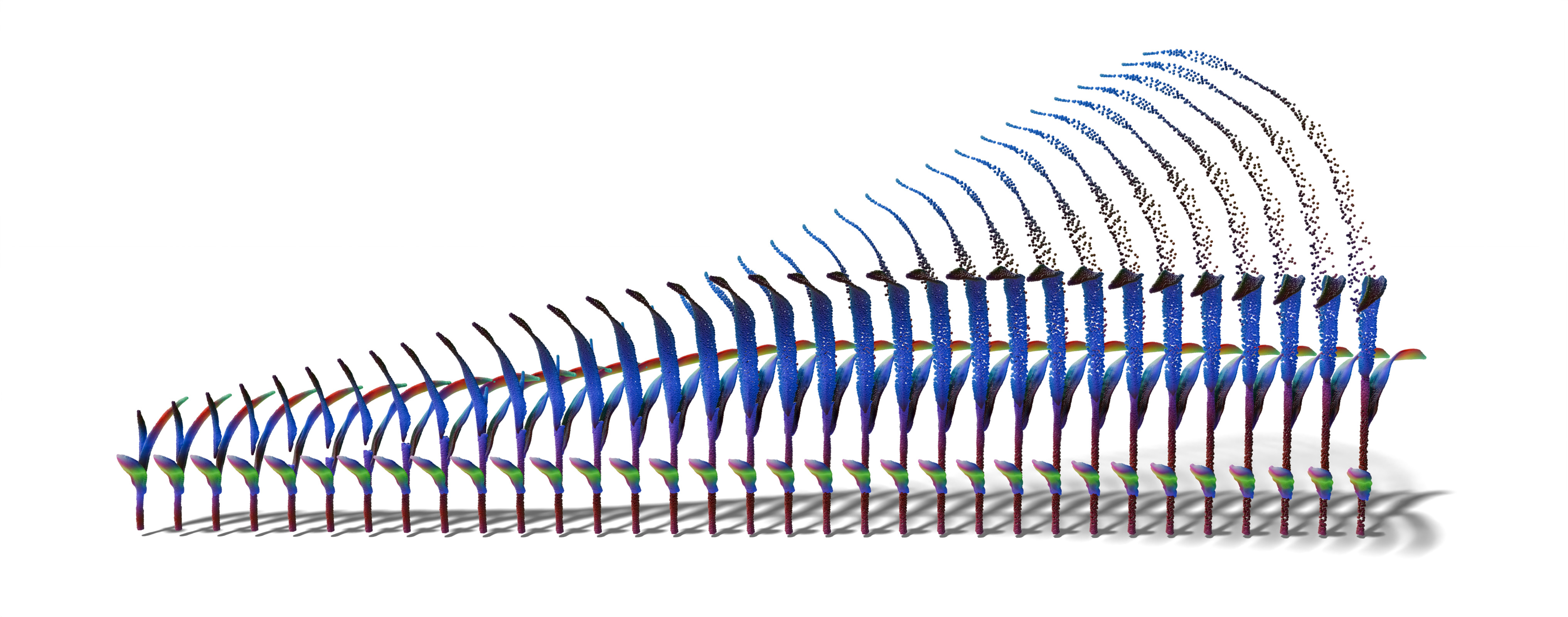}
    };
    \begin{scope}[x={(img.south east)},y={(img.north west)}]
    \draw[green!60!black, line width=0.8pt] (0.933,0.4) ellipse (0.03 and 0.05);
    \end{scope}
    
    \def\leftmargin{0.045}
    \def\rightmargin{0.9075}
    \foreach \i in {0,...,8} {
    \pgfmathsetmacro\t{\leftmargin + (\rightmargin-\leftmargin)*\i/8}
    \path ($(img.south west)!\t!(img.south east)$) coordinate (p);
    \ifodd\i
        \def\col{gray}\def\yshift{-0.25}
    \else
        \def\col{black}\def\yshift{0.0}
    \fi
    \node[\col,font=\scriptsize\bfseries,anchor=north] at ($(p)+(0,\yshift)$) {Day \i};
    }
    \end{tikzpicture}
    
    \end{tabular}
    };
    
    \begin{scope}[overlay,
        x={($(table.south east)-(table.south west)$)},
        y={($(table.north west)-(table.south west)$)},
        shift={(table.south west)}]
    \def\leftmargin{0.135}
    \def\rightmargin{0.895}
    \foreach \i in {0,...,8}{
        \pgfmathsetmacro\t{\leftmargin + (\rightmargin-\leftmargin)*\i/8}
        \ifodd\i
            \def\linecol{gray!20}
        \else
            \def\linecol{black!30}
        \fi
        \draw[\linecol, dashed, line width=0.2pt]
            (\t,0.1) -- (\t,0.8);
    }
    \end{scope}
    
    \end{tikzpicture}
        
    \vspace{-25pt}
    \caption{Point tracking of the densely interpolated \texttt{maize\_control\_plant2} sequence. \textbf{Black} labels indicate training frames, while \textcolor{gray}{\textbf{grey}} labels denote evaluation frames. Points are assigned a unique colour that remains consistent over time. Our method achieves equal or improved tracking consistency compared to the baselines.}
    \label{fig:interp_comparison_maize_control_plant2_tracking}

\end{figure}

We present dense interpolation results supporting \autoref{tab:interp_metrics} in the main paper in \autoref{fig:interp_comparison_maize_control_plant2} and \autoref{fig:interp_comparison_sorghum_control_plant2}.
Predicting the exact position and orientation of leaves at unseen frames is inherently ambiguous.
Nevertheless, our method produces plausible and smooth interpolations while providing improved organ joints, sharper leaf edges, and more consistent deformation fields than the baselines.
In addition, we show tracking results of points from the initial frame across the densely interpolated sequence in \autoref{fig:interp_comparison_maize_control_plant2_tracking}.
Additionally, we show quantitative interpolation results evaluating on per-organ sequences on \autoref{tab:interp_metrics_perorgan}.

\begin{table}[htbp]
\caption{\textbf{Per-organ interpolation results.}
Models are trained using every second timestep and evaluated on both observed (training) and interpolated frames.
DPF (per-organ) achieves the strongest interpolation performance overall, while our full model remains competitive among continuous neural-field approaches and performs particularly well on longer sequences.
The \textit{Mean} column reports the average across the plant sequences (each plant weighted equally).
All metrics are averaged over five independent runs.
}
\centering
\label{tab:interp_metrics_perorgan}
\resizebox{\textwidth}{!}{
\begin{tabular}{l cc cc cc cc cc cc cc cc cc}
\toprule
 & \multicolumn{2}{c}{MaC2} & \multicolumn{2}{c}{MaC3} & \multicolumn{2}{c}{SoC2} & \multicolumn{2}{c}{SoH2} & \multicolumn{2}{c}{TbC1} & \multicolumn{2}{c}{TbS3} & \multicolumn{2}{c}{Tm1H3} & \multicolumn{2}{c}{Tm1S1} & \multicolumn{2}{c}{\textit{Mean}} \\
\cmidrule(lr){2-3} \cmidrule(lr){4-5} \cmidrule(lr){6-7} \cmidrule(lr){8-9} \cmidrule(lr){10-11} \cmidrule(lr){12-13} \cmidrule(lr){14-15} \cmidrule(lr){16-17} \cmidrule(lr){18-19}
Method & CD $\downarrow$ & EPE $\downarrow$ & CD $\downarrow$ & EPE $\downarrow$ & CD $\downarrow$ & EPE $\downarrow$ & CD $\downarrow$ & EPE $\downarrow$ & CD $\downarrow$ & EPE $\downarrow$ & CD $\downarrow$ & EPE $\downarrow$ & CD $\downarrow$ & EPE $\downarrow$ & CD $\downarrow$ & EPE $\downarrow$ & CD $\downarrow$ & EPE $\downarrow$ \\
\midrule
\multicolumn{19}{c}{\textit{Training frames}} \\
COAP~\cite{mihajlovic_coap_2022} & 2.94 & 13.36 & 0.82 & \cellcolor{third} 3.31 & 2.77 & 12.56 & \cellcolor{third} 1.28 & \cellcolor{third} 0.84 & 0.71 & \cellcolor{second} 2.39 & 0.51 & 2.62 & 0.19 & 2.64 & 0.77 & \cellcolor{second} 1.69 & 1.25 & 4.93 \\
DPF (per-organ) & \cellcolor{third} 2.39 & \cellcolor{best} 2.08 & \cellcolor{second} 0.77 & \cellcolor{best} 0.72 & \cellcolor{second} 0.60 & \cellcolor{second} 0.96 & \cellcolor{second} 0.69 & 1.34 & \cellcolor{third} 0.65 & \cellcolor{best} 1.98 & \cellcolor{third} 0.49 & \cellcolor{best} 1.62 & \cellcolor{third} 0.17 & \cellcolor{second} 1.44 & \cellcolor{third} 0.63 & \cellcolor{third} 1.86 & \cellcolor{third} 0.80 & \cellcolor{best} 1.50 \\
Ours (per-organ MLPs) & \cellcolor{second} 2.34 & \cellcolor{third} 2.40 & \cellcolor{third} 0.79 & 3.58 & \cellcolor{third} 0.64 & \cellcolor{third} 1.81 & \cellcolor{second} 0.69 & \cellcolor{second} 0.69 & \cellcolor{second} 0.53 & \cellcolor{third} 3.44 & \cellcolor{second} 0.38 & \cellcolor{third} 1.86 & \cellcolor{second} 0.16 & \cellcolor{third} 1.65 & \cellcolor{second} 0.57 & 2.32 & \cellcolor{second} 0.76 & \cellcolor{third} 2.22 \\
\textbf{Ours (full)} & \cellcolor{best} 2.06 & \cellcolor{second} 2.24 & \cellcolor{best} 0.67 & \cellcolor{second} 1.39 & \cellcolor{best} 0.48 & \cellcolor{best} 0.76 & \cellcolor{best} 0.52 & \cellcolor{best} 0.31 & \cellcolor{best} 0.42 & 5.39 & \cellcolor{best} 0.32 & \cellcolor{second} 1.84 & \cellcolor{best} 0.14 & \cellcolor{best} 0.67 & \cellcolor{best} 0.53 & \cellcolor{best} 1.54 & \cellcolor{best} 0.64 & \cellcolor{second} 1.77 \\
\midrule
\multicolumn{19}{c}{\textit{Interpolated frames}} \\
COAP~\cite{mihajlovic_coap_2022} & \cellcolor{third} 21.00 & 16.15 & 10.54 & \cellcolor{third} 7.36 & 22.28 & 11.59 & 35.00 & 4.11 & 3.57 & \cellcolor{second} 3.97 & 2.96 & 2.77 & \cellcolor{best} 0.68 & 2.77 & \cellcolor{third} 1.48 & \cellcolor{second} 2.04 & \cellcolor{third} 12.19 & 6.34 \\
DPF (per-organ) & \cellcolor{best} 20.29 & \cellcolor{third} 8.77 & \cellcolor{best} 9.96 & \cellcolor{best} 5.44 & \cellcolor{best} 18.93 & \cellcolor{best} 6.64 & \cellcolor{third} 34.38 & \cellcolor{second} 3.50 & \cellcolor{third} 3.09 & \cellcolor{best} 3.23 & \cellcolor{third} 2.68 & \cellcolor{best} 2.21 & \cellcolor{third} 0.72 & \cellcolor{second} 1.69 & \cellcolor{second} 1.37 & \cellcolor{best} 1.98 & \cellcolor{second} 11.43 & \cellcolor{best} 4.18 \\
Ours (per-organ MLPs) & \cellcolor{second} 20.84 & \cellcolor{second} 8.69 & \cellcolor{second} 10.03 & 8.16 & \cellcolor{second} 18.99 & \cellcolor{third} 6.88 & \cellcolor{second} 34.20 & \cellcolor{best} 3.36 & \cellcolor{best} 2.71 & \cellcolor{third} 4.67 & \cellcolor{best} 2.49 & \cellcolor{third} 2.43 & 0.73 & \cellcolor{third} 1.95 & \cellcolor{second} 1.37 & 3.16 & \cellcolor{best} 11.42 & \cellcolor{third} 4.91 \\
\textbf{Ours (full)} & 21.10 & \cellcolor{best} 8.35 & \cellcolor{third} 10.13 & \cellcolor{second} 5.82 & \cellcolor{third} 19.68 & \cellcolor{second} 6.66 & \cellcolor{best} 32.97 & \cellcolor{third} 3.52 & \cellcolor{second} 2.78 & 6.54 & \cellcolor{second} 2.67 & \cellcolor{second} 2.28 & \cellcolor{second} 0.71 & \cellcolor{best} 1.22 & \cellcolor{best} 1.31 & \cellcolor{third} 2.30 & \cellcolor{best} 11.42 & \cellcolor{second} 4.59 \\
\bottomrule
\end{tabular}
}
\end{table}

\section*{Qualitative extrapolation results}
\begin{figure}[t]
\centering
\begin{tikzpicture}
  \node[anchor=south west, inner sep=0] (img) {\includegraphics[width=\linewidth]{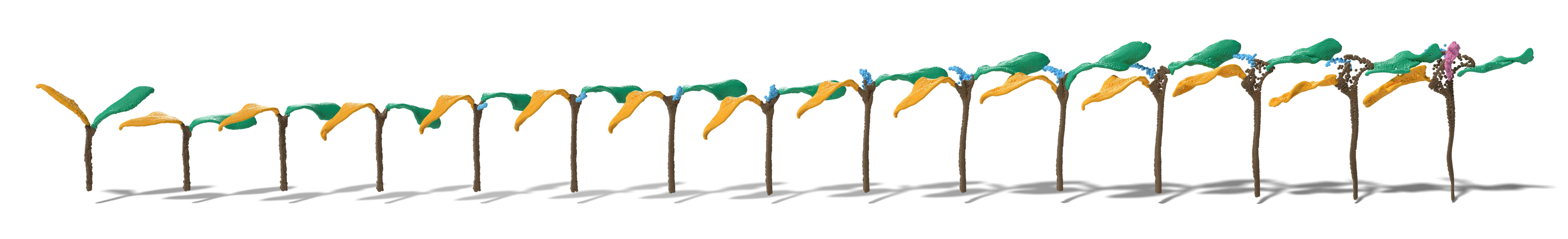}};
  \begin{scope}[x={(img.south east)},y={(img.north west)}]
    \draw[eccvblue, very thick, dashed, rounded corners=3pt] (0.76,-0.08) rectangle (0.99,1.02);
    \node[eccvblue, anchor=south, font=\scriptsize\bfseries] at (0.875,-0.06) {Extrapolation};
  \end{scope}
\end{tikzpicture}
\caption{\textbf{Qualitative extrapolation on \texttt{tomato1\_heat\_plant3}.} GrowFields prediction, trained on all but the last three timesteps and rolled out from $t_0$ across the full sequence (left to right); colours denote organ identities, and the dashed box marks the three held-out frames evaluated for extrapolation. On this longer sequence the model produces plausible continued growth into the held-out frames, consistent with its low extrapolation error on Tm1H3 (\autoref{tab:extrapolation}).}
\label{fig:extrapolation_qual_tomato}
\end{figure}

We provide an additional extrapolation example on a longer sequence, \texttt{tomato1\_\allowbreak heat\_\allowbreak plant3} (Tm1H3, 16 timesteps), in \autoref{fig:extrapolation_qual_tomato}, complementing \autoref{fig:extrapolation_qual} in the main paper.

\section*{Robustness to noise and missing data}
To assess robustness to imperfect inputs, we perturb the test sequences in two ways and re-evaluate GrowFields.
First, we add Gaussian noise with $\sigma \in \{0.01,\allowbreak 0.05\}$\,mm to all points.
Second, we create \emph{holes} by removing five local neighbourhoods of $K\in\{15,25\}$ training points per leaf, so that prediction must occur in regions with no nearby training points.
As shown in \autoref{tab:robustness}, compared to the unperturbed setting (mean CD~0.86 / EPE~1.47, Table~1 in the main paper), the model degrades gracefully under realistic noise (mean CD~1.08 at $\sigma\!=\!0.01$~mm) and remains stable under missing data.
Crucially, the holes experiment confirms that GrowFields learns genuine growth dynamics rather than exploiting nearest-neighbour proximity between train and test points.

\begin{table}[htbp]
\centering
\caption{\textbf{Robustness to input perturbations.} Mean per-sequence CD~$\downarrow$ / EPE~$\downarrow$ under additive Gaussian noise and removal of local point neighbourhoods (5 holes of $K$ points). GrowFields degrades gracefully and remains effective even when test points have no nearby training neighbours. All metrics are averaged over five independent runs.}
\label{tab:robustness}
\resizebox{\textwidth}{!}{
\begin{tabular}{l cc cc cc cc cc cc cc cc cc}
\toprule
 & \multicolumn{2}{c}{MaC2} & \multicolumn{2}{c}{MaC3} & \multicolumn{2}{c}{SoC2} & \multicolumn{2}{c}{SoH2} & \multicolumn{2}{c}{TbC1} & \multicolumn{2}{c}{TbS3} & \multicolumn{2}{c}{Tm1H3} & \multicolumn{2}{c}{Tm1S1} & \multicolumn{2}{c}{\textit{Mean}} \\
\cmidrule(lr){2-3} \cmidrule(lr){4-5} \cmidrule(lr){6-7} \cmidrule(lr){8-9} \cmidrule(lr){10-11} \cmidrule(lr){12-13} \cmidrule(lr){14-15} \cmidrule(lr){16-17} \cmidrule(lr){18-19}
Perturbation & CD $\downarrow$ & EPE $\downarrow$ & CD $\downarrow$ & EPE $\downarrow$ & CD $\downarrow$ & EPE $\downarrow$ & CD $\downarrow$ & EPE $\downarrow$ & CD $\downarrow$ & EPE $\downarrow$ & CD $\downarrow$ & EPE $\downarrow$ & CD $\downarrow$ & EPE $\downarrow$ & CD $\downarrow$ & EPE $\downarrow$ & CD $\downarrow$ & EPE $\downarrow$ \\
\midrule
Noise $\sigma\!=\!0.01$ & 3.03 & 2.74 & 2.02 & 4.33 & 1.04 & 1.98 & 0.88 & 1.38 & 0.63 & 2.25 & 0.54 & 1.57 & 0.13 & 0.96 & 0.34 & 1.34 & 1.08 & 2.07 \\
Noise $\sigma\!=\!0.05$ & 15.46 & 8.76 & 10.50 & 11.05 & 6.79 & 8.00 & 3.40 & 3.99 & 3.13 & 2.42 & 1.43 & 2.58 & 0.27 & 1.11 & 0.96 & 1.88 & 5.24 & 4.97 \\
\midrule
Holes $K\!=\!15$ & 2.79 & 2.50 & 1.61 & 2.05 & 0.85 & 1.84 & 0.85 & 1.40 & 0.65 & 2.50 & 0.47 & 2.06 & 0.13 & 1.05 & 0.34 & 1.45 & 0.96 & 1.86 \\
Holes $K\!=\!25$ & 3.01 & 2.99 & 1.81 & 2.04 & 1.03 & 2.03 & 0.96 & 1.74 & 0.69 & 2.76 & 0.52 & 1.79 & 0.14 & 0.93 & 0.37 & 1.39 & 1.07 & 1.96 \\
\bottomrule
\end{tabular}
}
\end{table}

\vspace{-4mm}

\section*{Applications}
\begin{figure}
    \centering
    \includegraphics[width=0.8\linewidth]{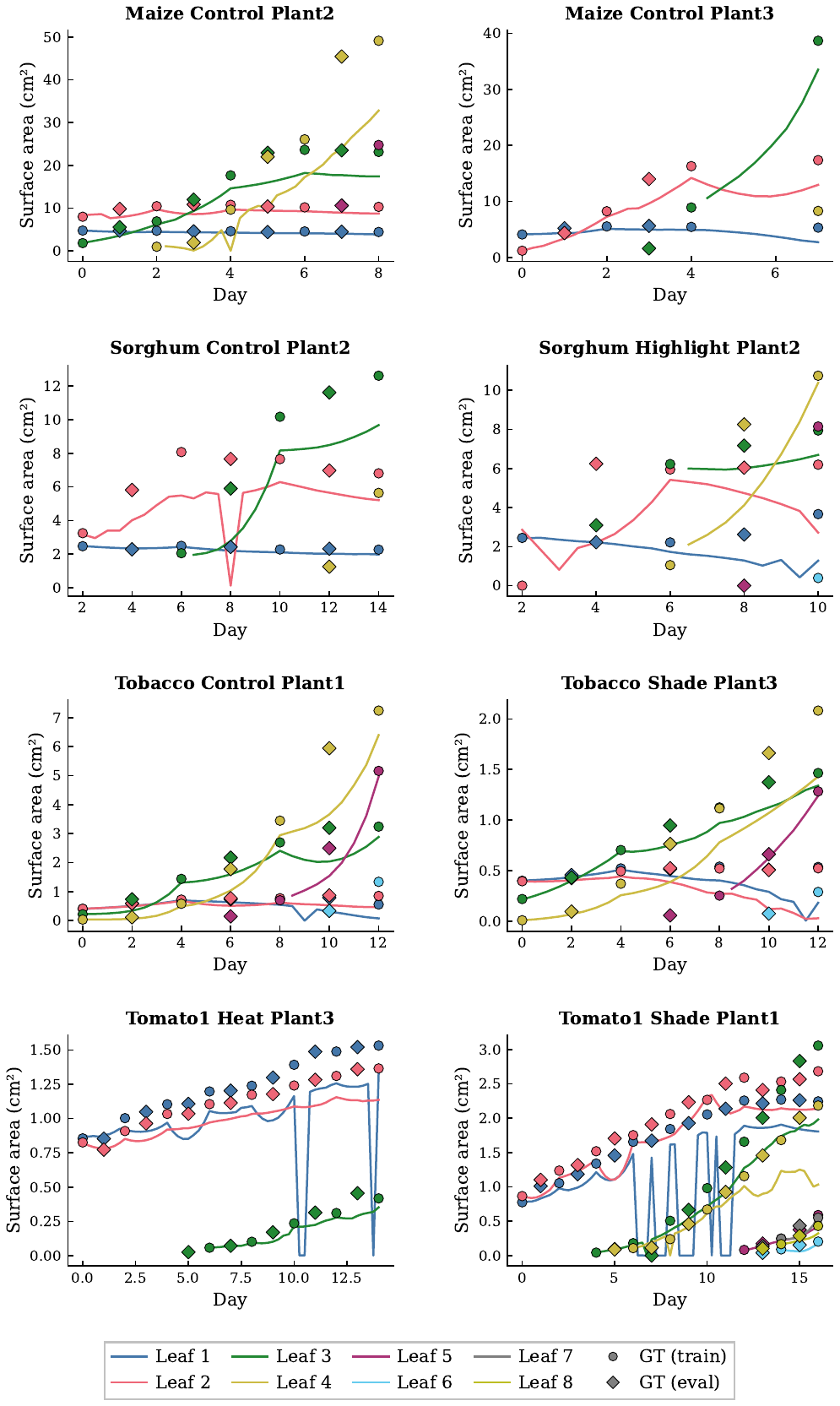}
    \caption{Estimated leaf surface area over time for all test sequences. We reconstruct surfaces using the Ball Pivoting algorithm~\cite{Bernardini1999_ball_pivoting} and compute leaf surface area from the resulting meshes. The estimated areas from interpolated frames are compared to those of ground-truth frames. Our method closely follows the ground-truth trend while providing smooth temporal estimates. Failures primarily occur in regions with a low number of points per organ.}
    \label{fig:Leaf_area}
\end{figure}

Given the smooth and well-distributed growth fields produced by our method, we can estimate morphological plant and organ traits during interpolation.
As an example, we compute the leaf surface area during interpolation and compare it to the estimated leaf area of the ground-truth frames in \autoref{fig:Leaf_area}.
For both cases the Ball Pivoting algorithm~\cite{Bernardini1999_ball_pivoting} is used.

\end{document}